\documentclass[10pt,twocolumn,letterpaper]{article}

\usepackage[pagenumbers]{styleguide} % To force page numbers, e.g. for an arXiv version

\usepackage{url}
\usepackage{svg}
\usepackage[separate-uncertainty = true,multi-part-units=single]{siunitx}
\usepackage{booktabs}
\usepackage{subcaption} %  for subfigures environments 
\usepackage{enumitem}
\usepackage{amssymb}
\usepackage{amsmath}
\usepackage{multirow}
\usepackage{xspace}
\usepackage{wrapfig}
\usepackage{tablefootnote}
\usepackage{xcolor}%

\usepackage{float}
\usepackage{makecell}
\usepackage{wasysym}
\usepackage{pifont}
\usepackage{tikz}
\usepackage{graphicx}
\usepackage{cuted} %main package: to insert the image

\usepackage[pagebackref,breaklinks,colorlinks,allcolors=iccvblue]{hyperref}

\newcommand{\frameworkshort}{ReCL\xspace}
\newcommand{\frameworklong}{\emph{\textbf{Re}construction for \textbf{C}ontinual \textbf{L}earning}\xspace}
\definecolor{iccvblue}{rgb}{0.21,0.49,0.74}

\title{Slowing Down Forgetting in Continual Learning}

\author{Pascal Janetzky\\
Bosch Center for Artificial Intelligence\\
Munich Center for Maschine Learning\\
{\tt\small pascal.janetzky@bosch.com}
\and
Tobias Schlagenhauf\\
Bosch Center for Artificial Intelligence\\
{\tt\small tobias.schlagenhauf@bosch.com}
\and
Stefan Feuerriegel\\
Munich Center for Machine Learning \& LMU Munich\\
{\tt\small feuerriegel@lmu.de}
}

\begin{document}
\maketitle

\begin{abstract}
A common challenge in continual learning (CL) is catastrophic forgetting, where the performance on old tasks drops after new, additional tasks are learned. In this paper, we propose a novel framework called \frameworkshort to slow down forgetting in CL. Our framework exploits an implicit bias of gradient-based neural networks due to which these converge to margin maximization points. Such convergence points allow us to reconstruct old data from previous tasks, which we then combine with the current training data. Our framework is flexible and can be applied on top of existing, state-of-the-art CL methods. We further demonstrate the performance gain from our framework across a large series of experiments, including two challenging CL scenarios (class incremental and domain incremental learning), different datasets (MNIST, CIFAR10, TinyImagenet), and different network architectures. Across all experiments, we find large performance gains through \frameworkshort. To the best of our knowledge, our framework is the first to address catastrophic forgetting by leveraging models in CL as their own memory buffers.
\end{abstract}

\section{Introduction}

{Continual learning (CL)} is a paradigm in machine learning where models are trained continuously to adapt to new data without forgetting previously learned information \citep{parisi2019continual,wang2024comprehensive}. This is relevant for real-world applications such as autonomous driving \citep{shaheen2022continual}, medical devices \citep{vokinger2021continual}, predictive maintenance \citep{hurtado2023continual}, and robotics \citep{thrun1995lifelong}. In these applications, the entire data distribution cannot be sampled prior to model training and (fully) storing the arriving data is not possible. 

The main challenge of CL is \emph{catastrophic forgetting} \citep{mccloskey1989catastrophic}. Catastrophic forgetting refers to problems where the performance on old data drops as new data is learned. Catastrophic forgetting typically happens because prior knowledge is encoded in the combination of the model parameters, yet updating these model parameters during training on new data leads to a change in knowledge and can thus cause the model to forget what it has learned before.

To slow down forgetting, numerous methods have been proposed \citep{rusu2016progressive,kirkpatrick2017overcoming,li2017learning,lopez2017gradient,mallya2018packnet,rolnick2019experience,yoon2020scalable,deng2021flattening,wang2021ordisco,kang2022forget,pmlr-v235-malagon24a}. Existing methods can be categorized into two main groups \citep{saha2021gradient,elsayed2024addressing,wang2024comprehensive}: (1)~\emph{Memory-based} methods rely on external memory to either store samples from old tasks \citep[e.g.,][]{rebuffi2017icarl,rolnick2019experience} or store separately trained generative networks for generating old data on demand \citep[e.g.,][]{shin2017continual,mundt2022unified}. However, this is often not practical due to the fact that CL is often used because streaming data becomes too large and cannot be stored. (2)~\emph{Memory-free methods} regularize parameter updates, which may help to slow down forgetting at the cost of learning new concepts more slowly \citep[e.g.,][]{kirkpatrick2017overcoming,saha2021gradient,elsayed2024addressing}. Here, we contribute a different, orthogonal strategy in which we propose to leverage the trained model as its own memory buffer. 

In this paper, we develop a novel framework, called {\frameworklong} (\frameworkshort), to slow down forgetting in CL. Different from existing works, our framework leverages the implicit bias of gradient-based neural network training, which causes convergence to margin maximization points. Such convergence points essentially memorize old training data in the model weights, which allows us then to perform a dataset reconstruction attack. We then integrate this reconstruction attack into CL, where we minimize a data reconstruction loss to recover the training samples of old tasks. Crucially, our framework is flexible and can be used on top of existing CL methods to slow down forgetting and thus improve performance. 

We evaluate the effectiveness of our framework across a wide range of experiments. (1)~We show the performance gains from our framework across two challenging CL scenarios: class incremental learning and domain incremental learning. (2)~We compare the performance across different and widely-used datasets, namely MNIST~\citep{lecun1998gradient}, CIFAR10~\citep{alex2009learning}, and TinyImageNet~\citep{le2015tiny} (3)~We compare both multi-layer perceptrons (MLPs) and convolutional neural networks (CNNs), finding that consistent performance gains can be achieved across different network architectures. (4)~Lastly, we demonstrate the flexibility of our framework by combining our \frameworkshort frameworks with several state-of-the-art CL methods, namely, {EWC} \citep{kirkpatrick2017overcoming}, {ER} \citep{rolnick2019experience}, {UPGD} \citep{elsayed2024addressing}, {AGEM} \citep{lopez2017gradient}, {LwF} \cite{li2017learning}, and {GPM} \citep{saha2021gradient}. Here, we find  that our framework improves over the vanilla version of the CL methods and can slow down forgetting further. Across all experiments, we find a consistent performance gain from using our \frameworkshort framework. 

\textbf{Contributions:}\footnote{Code: \href{https://anonymous.4open.science/r/slower-forgetting/}{https://anonymous.4open.science/r/slower-forgetting/}; upon acceptance, we will move the code to a public repository.}  
(\textbf{1})~ We present a novel CL framework called \frameworkshort to slow down forgetting in CL. To the best of our knowledge, ours is the first method to slow down catastrophic forgetting in which trained models are leveraged as their own memory buffers. (\textbf{2})~We propose to leverage the implicit bias of margin maximization points to create training data in CL. (\textbf{3})~We demonstrate that our \frameworkshort is flexible and that it consistently slows down forgetting.

\section{Related Work}
\label{sec:related_work}
\textbf{Continual learning:} CL is a machine learning paradigm where models are trained on newly-arriving data that cannot be stored \citep{wang2024comprehensive}. In the literature, the following two scenarios are considered especially challenging\citep{hsu2018re,van2019three,van2022three} and are thus our foucs. (1)~In \textbf{class incremental learning} (CIL), a model is trained on new classes that arrive sequentially. Here, each task has a unique label space, and the model's output layer is expanded to incorporate the new classes. (2)~In \textbf{domain incremental learning} (DIL), the size of the output layer is fixed, and each task uses the same label space. However, the underlying data that correspond to the labels are no longer fixed but can change over time (e.g., indoor images of cats $\rightarrow$ outdoor images of cats). Later, we apply our framework to both CL scenarios (CIL and DIL) and show that it achieves consistent performance gains. 

\textbf{Forgetting:} A main challenge in CL is \emph{forgetting}, also named catastrophic forgetting or catastrophic inference \citep{mccloskey1989catastrophic,wang2024comprehensive}. The problem in forgetting is that, as the model parameters are updated in response to new data, important weights for older tasks are altered and may cause a drop in performance for older tasks~\citep{mallya2018packnet,saha2021gradient}.  Note that forgetting is a problematic issue in all CL scenarios. \citep{wang2024comprehensive}. One may think that a na{\"i}ve workaround is to restrict the magnitude of how much model parameters can be modified during training; yet, this also restricts how new tasks are learned and essentially leads to a stability-plasticity tradeoff \citep{wang2024comprehensive}. Hence, such a na{\"i}ve workaround is undesired.

To reduce forgetting, several methods have been proposed, which can be grouped into two major categories (see \Cref{app:extendend_lit} for a detailed review). (1)~\emph{Memory-based} methods rely on external memory to either directly store samples from old tasks \citep[e.g.,][]{rebuffi2017icarl,rolnick2019experience} or store separately trained generative networks for generating old data on demand \citep[e.g.,][]{shin2017continual,mundt2022unified}. A prominent example is experience replay (\textbf{ER}) where samples of old data are replayed from a small memory buffer \citep{rolnick2019experience}. Another example is \textbf{AGEM} \citep{chaudhry2018efficient}, which stores sold samples to alter the gradient update of subsequent tasks. (2)~\emph{Memory-free} approaches do not store old data and can be further grouped into two subcategories. (i)~One stream adopts architecture-based approaches, which alter the model architecture to more flexibly accommodate new tasks (e.g., by adding task-specific modules \citep{rusu2016progressive,yoon2018lifelong,ke2020continual}. Oftentimes, these approaches avoid forgetting altogether (by freezing old modules), but at the cost of exponentially growing network sizes. Such approaches usually require task identities and are thus mainly relevant for specific CL scenarios such as task-incremental learning, because of which methods from this subcategory are generally \emph{not} applicable as baselines. (ii)~Another stream of approaches regularizes parameter updates so that weights are updated based on their importance for old tasks \citep{kirkpatrick2017overcoming,zenke2017continual,saha2021gradient}. Yet, this typically also interferes with learning from new data. Prominent examples are: Elastic weight consolidation (\textbf{EWC}) \citep{kirkpatrick2017overcoming}, which uses the Fisher information matrix to assess parameter importance and to regularize gradient updates to parameters. Learning without Forgetting (\textbf{LwF}) \citep{li2017learning} penalizes changes to a models output distribution, compared to the beginning of a task, when data from the current task is passed through. (iii) Yet another stream modifies the gradient update through projection methods. An example is gradient projection memory (\textbf{GPM}) \citep{saha2021gradient}, which computes gradient subspaces that are already occupied by previous tasks. It then uses gradient-preconditioning matrices to guide new gradient updates to be orthogonal. A second example is utility-perturbed gradient descent (\textbf{UPGD}) \citep{elsayed2024addressing}, which measures parameter importance by computing the performance difference between unmodified and perturbed weights, and uses this per-parameter utility to control plasticity by scaling a noised gradient.

\textbf{Difference to generative replay:} Our proposed framework is different to both memory-based and memory-free approaches. We do \textbf{\emph{not}} store data of old tasks, and we also do \textbf{\emph{not}} train or keep additional generative networks. Specifically, generative replay uses \textit{separate} generators, which require \textit{additional} training, additional design choices, and an additional labeling network. Even in cases where classifier and generator are merged, such as \cite{van2020brain} (they use a VAE and apply a reconstruction loss to internal hidden representations and optionally use SI \cite{zenke2017continual} to regularize parameter updates), architectural modifications are necessary. \frameworkshort has four advantages: \textbf{(1)}~Our \frameworkshort does \underline{not} require a separate generator. Instead, it uses a custom reconstruction loss to ``sample'' old data from the classifier network. \textbf{(2)}~The loss is \underline{not} based on reconstructed or internal representations. Instead, it optimizes data points so that their derivatives linearly approximate the network weights. \textbf{(3)}~We do \underline{not} need to label the reconstructed data, as \emph{the data are already labeled}. \textbf{(4)}~Lastly, we do \underline{not} meddle with the network architecture, and also do \underline{not} alter the learning objective, as this typically interferes with learning new tasks. Rather, we keep the network as-is and combine the reconstructed training data with the current task's data. Finally, our proposed \frameworkshort is flexible and, as we show later, can be combined with any of the mentioned baselines, where our \frameworkshort leads to further performance improvements.  

\textbf{Dataset reconstruction:} The task of dataset reconstruction is related to the idea of memorization in neural networks \citep{carlini2019secret,brown2021memorization,buzaglo2024deconstructing}. Specifically, dataset reconstruction builds upon the implicit bias of gradient-based neural network training, which causes neural networks to generalize despite fitting to training-data specific patterns \citep{zhang2021understanding}. More technically, \citet{lyu2019gradient} and \citet{ji2020directional} show that, despite the freedom to converge to any point during training, homogeneous neural networks are essentially biased toward converging against margin maximization points. Under such a convergence condition, \citet{haim2022reconstructing} even show that the \emph{entire} training data can be recovered from pretrained binary classifiers without access to any reference data. \citet{buzaglo2024deconstructing} extend the underlying theory to multi-class networks and general loss functions. 

The above-mentioned implicit bias has been studied for regression tasks \citep{amid2020reparameterizing,azulay2021implicit,nacson2022implicit}, classification tasks \citep{gunasekar2018characterizing,moroshko2020implicit,nacson2023implicit}, and also different optimization techniques \citep{blanc2020implicit,damian2021label,li2021happens,wang2021implicit}. However, exploiting this implicit bias  has \textbf{\emph{not}} yet been studied in CL. Hence, our framework is orthogonal to the above research directions as we leverage dataset reconstruction attacks for CL, which is our novelty. 

\textbf{Research gap:} To the best of our knowledge, the implicit bias of gradient-trained neural networks to converge to margin maximization points and thus to memorize training data has been overlooked in CL. Here, we propose to exploit this implicit bias, which enables us to reconstruct training data from machine learning models and which then allows us to slow down forgetting in CL. For this, we develop a novel framework called \frameworklong{} (\frameworkshort). 

\section{Problem Statement}\label{sec:problem_definition}
We consider the following, standard setup for CL, where data (e.g., images) arrive sequentially and where a model is updated continually to learn from a series of tasks \citep{rusu2016progressive,neyshabur2017exploring,yoon2020scalable,saha2021gradient}.

\textbf{Tasks:} Each task $\tau \in \{1,2,\dots,T\}$ has its own data (sub-)set $\mathcal{D}_\tau = { (x_{i}^{\tau}, y_{i}^{\tau}) }_{i=1}^{n_{\tau}} \subseteq \mathbb{R}^d \times C $ with features $x_i$ and corresponding labels $y_i$ from classes $C = \{1, \ldots, C\}$. We use the term `task' or `time' interchangeably when referring to $\tau$. Note that the data cannot be stored, meaning that, at time $\tau$, the current dataset $\mathcal{D}_{\tau}$ with $(\mathbf{x}^{\tau}, \mathbf{y}^{\tau})$ is only available for training the current task but cannot be stored. For each task, we have separate train and test split given by $\mathcal{D}_{\tau}^{tr}$ and $\mathcal{D}_{\tau}^{te}$, respectively.

\textbf{Network $\Phi$:} We train a homogenous\footnote{See \Cref{app:homogenous_definition} for a formal definition.} neural network $\Phi(\theta,\cdot)$, which consists of a feature extraction part $\phi_f: \mathbb{R}^d \rightarrow \mathbb{R}^f$ and classification head $\phi_h: \mathbb{R}^f\mapsto C$ (typically, a single dense layer). Together, both $\phi_f$ and $\phi_h$ implement an input-output mapping $\Phi(\theta,\cdot): (\phi_f \circ \phi_h)= \mathbb{R}^d \mapsto C$. We denote the network and its parameters at time $\tau$ by $\Phi_\tau$. Note that the class of homogenous neural networks is very broad. For example, it includes all vanilla MLPs and CNNs without components introducing ``jumps'' into the forward pass (such as, e.g., bias vectors or dropout~\citep{srivastava2014dropout}).

\textbf{Scenarios:} We consider the following two common scenarios in CL \citep[for a detailed overview, see][]{hsu2018re,van2022three}. The scenarios vary in how the classification head $\phi_h$ is used:
\begin{enumerate}[leftmargin=+0.5cm]
    \item \textbf{CIL:} Each task $\tau$ introduces a new set of classes $C_{\tau} \subset C$. Notably, the label space of any two tasks does not overlap ($C_{\tau} \cap C_{\tau'} = \emptyset, \forall\, \tau \neq \tau'$). At each task, the shared $\phi_h$ is expanded to incorporate the novel classes, while copying the learned parameters from the old output head. Task-identity is only available at train-time.
    \item \textbf{DIL:} A single shared head is used for each task, but the architecture is not modified upon the arrival of a new $\tau$. Rather, each task introduces new data for already known classes, so that the existing head is simply trained further. Task-identity is available at train-time only.
\end{enumerate}
These scenarios vary in terms of difficulty, and CIL is often considered to be the most challenging one, as novel concepts need to be learned without forgetting old concepts \citep[cf.][]{hsu2018re,van2022three}.
\begin{figure*}[!htbp]
  \centering
  \includegraphics[]{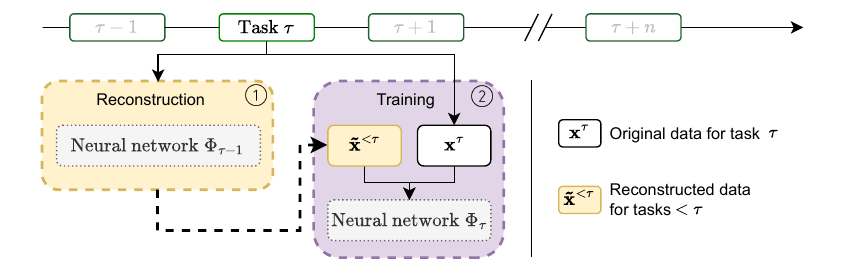}
  \caption{\textbf{Overview of our \frameworkshort framework}. Upon the arrival of a new task $\tau$, we reconstruct samples from all previous tasks from the network $\Phi_{\tau-1}$ by minimizing \Cref{eq:loss_full}. The reconstructed samples are then combined with the current task's data. Finally, the model is trained on the combined dataset, yielding the new network $\Phi_{\tau}$, from which data will be reconstructed in $\tau+1$.}\label{fig:framework}
\end{figure*}
\textbf{Metrics:} We measure forgetting using the following standard metrics: (i)~\emph{average accuracy} (ACC) \citep{lopez2017gradient} and (ii)~\emph{backwards transfer} (BWT) \citep{lopez2017gradient}. ACC captures the accuracy on all tasks up to task $T$ via
\begin{equation}
    \text{ACC} = \frac{1}{T} \sum_{\tau=1}^{T} \mathit{acc}_{T,\tau} ,
\end{equation}
where $\mathit{acc}_{T,\tau}$ is the classification accuracy on task $\tau$ after training on task $T$, meaning it gives the averaged test set performance over all tasks. BWT measures the performance loss by contrasting the original task performance with performance after training on subsequent tasks, which thus reflects forgetting. Formally, 
\begin{equation}
    \text{BWT} = \frac{1}{T-1} \sum_{\tau=1}^{T-1} \mathit{acc}_{T,\tau} - \mathit{acc}_{\tau,\tau} ,
\end{equation}
where $\mathit{acc}_{T,\tau}$ is the accuracy on a previous task $\tau$ after training on task $T$, and $\mathit{acc}_{\tau,\tau}$ is the performance on $\tau$ directly after training on $\tau$. For BWT, negative numbers imply forgetting, while positive numbers indicate a retrospective accuracy improvement (which is desired).

\textbf{Our objective:} We aim to \textbf{\emph{slow down forgetting}} in a continually trained neural network $\Phi$ that is subject to continual learning. For this, we present a framework where we exploit the implicit bias of gradient-based neural network training towards margin maximization points.

\section{Proposed \frameworkshort Framework}
In this section, we describe the components of our proposed framework for slowing down forgetting by dynamically extracting the training data of old tasks. 

\textbf{Overview:} An overview of our \frameworkshort framework is shown in \Cref{fig:framework}. Our framework performs the following two steps upon arrival of a new task $\tau$. \underline{Step~1:} \textbf{Reconstructing} the training data from $\theta_{\tau-1}$, thus generating the reconstructed data $\Tilde{\mathcal{D}}_\tau = (\mathbf{\Tilde{x}}^{<\tau},\mathbf{\Tilde{y}}^{<\tau})$. \underline{Step~2:} \textbf{Combining} the reconstructed data of previous tasks with the current data $(\mathbf{x}^{\tau}, \mathbf{y}^{\tau})$ and then performing further training of $\Phi_{\tau}(\theta,\cdot)$. Note that both steps are independent of the underlying CL methods, meaning that our framework can be applied on top of \emph{\textbf{any}} existing CL method to slow down forgetting.

\textbf{Leveraging the implicit bias of convergence to margin maximization points:} Our \frameworkshort framework exploits the implicit bias of gradient-based neural network training, which causes neural network weights to converge to margin maximization points \cite{lyu2019gradient,ji2020directional}. These convergence points are characterized by several conditions \citep{lyu2019gradient,yu2023generator,buzaglo2024deconstructing} which, among others, state that the network weights $\theta$ can be approximated by a linear combination of the network's gradients on data points $x_i$. We utilize this finding and optimize \emph{randomly initialized} $x_i$ to closely resemble the training data of previous tasks. We later achieve this by minimizing a tailored reconstruction loss. 

\subsection{Step 1: Data Reconstruction for CL}\label{sec:step_data_reconstruction}

For data reconstruction, we perform two sub-steps: (1)~We first \emph{randomly} initialize a set of $m$ reconstruction candidates $x_i$, and (2)~we then optimize the candidates to satisfy our objective of reconstructing the training data from old tasks. Both are as follows:

For sub-step (1), we simply initialize $m$ candidates $x_i \sim \mathcal{N}(0, \sigma_x^2 \mathbb{I})$ and set \emph{fixed} labels evenly split across the number of classes.\footnote{The label information can be derived from the output layer of the trained neural network.} That is, $y_i \gets U(C)$.

For sub-step (2), we then optimize the $x_i$ to resemble old data by minimizing a combination of three losses over $n_\text{rec}$ reconstruction epochs. The losses are as follows: (i)~the \emph{reconstruction loss} $L_\text{rec}$, (ii)~the \emph{lambda loss} $L_{\lambda}$, and (iii)~the \emph{prior loss} $L_{\text{prior}}$. Here, the reconstruction loss in (i) denotes our central objective: it optimizes the candidates to reconstruct old data. However, to ensure the theoretical foundation of the implicit bias, we must ensure that the network weights can be approximated by a linear combination, for which we need the lambda loss in (ii), which enforces a constraint on the coefficients of the linear combination. To incorporate prior domain knowledge about the reconstructed data (such as having knowledge about its value range), we need the prior loss in (iii). It forces the candidates to lie within (normalized) value ranges. The formal definitions of the different losses are below:\footnote{We provide further details on the reconstruction process in \Cref{app:ds_recon_background}.}

\textbf{(i) Reconstruction loss:} The \emph{reconstruction loss} $L_\text{rec}$ optimizes the $m$ randomly initialized candidate samples $x_i$. Inspired by \citet{buzaglo2024deconstructing}, we define it as
\begin{multline}
    L_{\text{rec}}(x_1, \ldots, x_m, \lambda_1, \ldots, \lambda_m) =\\ \left\| \theta_{\tau-1} - \sum_{i=1}^{m} \lambda_i \nabla_{\theta_{\tau-1}} \Phi_{\tau-1}(\theta_{\tau-1}; x_i ) \right\|_2^2\text{,}\label{eq:loss_rec}
\end{multline}
where $ \left\|\cdot\right\|_2^2$ denotes the squared $\text{L}_2$ norm (i.e., squared Euclidean distance), $x_i$ are reconstruction candidates, $\lambda_i$ are learnable scaling coefficients. The number of reconstruction candidates, $m$, is a hyperparameter.

\textbf{(ii) Lambda loss:} The \emph{lambda loss} $L_{\lambda}$ constrains the scaling coefficients $\lambda_i$. It is defined as
\begin{equation}
    L_{\lambda} = \sum_{i=1}^{m} \text{max} \left\{ -\lambda_i, -\lambda_{\text{min}}\right\}\text{,}
\end{equation}
where $\lambda_{min}$ is a hyperparameter. Note that the $\lambda_i$ are not part of the old task's training data, but are necessary for the underlying theory to hold \citep{haim2022reconstructing} and thus to solve the reconstruction problem defined by \Cref{eq:loss_rec}. We provide further details in \Cref{sec:app_ds_recon}.

\textbf{(iii) Prior loss:} The \emph{prior loss} $L_{\text{prior}}$ constrains the value range of the (reconstructed) images to lie within normalized range $-1$ to $1$. Generally, this loss is used to encode prior background/domain knowledge about the reconstruction data. Here, we follow \citep{haim2022reconstructing} and use it to restrict that the values of the reconstructed data are in a certain range. Later, in our experiments, we work with images and thus enforce normalization to $[-1, 1]$ by
\begin{equation}
    L_\text{prior}=\sum_{i=1}^m \sum_{k=1}^d \max \{\max \{x_{i,k}-1,0 \}, \max \{ -x_{i,k}-1,0 \} \},
\end{equation}
where $d$ is the dimensionality of the features $x_i$.

\textbf{Overall loss:} For each task, we minimize the following overall loss 
\begin{equation}
    L_\text{full} = L_\text{rec} + L_{\lambda} + L_{\text{prior}}\label{eq:loss_full} 
\end{equation}
using stochastic gradient descend (SGD) over $n_\text{rec}$ reconstruction epochs. This yields the $\Tilde{\mathcal{D}}_\tau = (\mathbf{\Tilde{x}}^{<\tau},\mathbf{\Tilde{y}}^{<\tau})$. 

\subsection{Step 2: Combining the Reconstructed Data for Training}
\label{sec:step_training}

Step~2 now leverages the reconstructed data, $\Tilde{\mathcal{D}}_\tau$, when training for the new task $\tau$. We set the number of samples to reconstruct to $m=\sum_{t=1}^{\tau-1}n^t$ and initialize the fixed labels $\mathbf{\Tilde{y}}^{<\tau}$ evenly across all classes. We then combine $\Tilde{\mathcal{D}}_\tau$ with the current training data $\mathcal{D_{\tau}}^{tr}$ and train $\Phi_{\tau}$ on the combined dataset.

\subsection{In-Training Optimization of Reconstruction Hyperparameters}\label{sec:hyperparameter_tuning_strategies}

Our \frameworkshort consists of six hyperparameters, which are primarily used in \Cref{eq:loss_rec} to \Cref{eq:loss_full}. The hyperparameters are: (i)~$\lambda_{\text{min}}$ constrains the scaling effect of $\lambda_i$; (ii)~$\sigma_x$ controls the initialization of the $x_i$ from a Gaussian distribution; (iii)~$\text{lr}_{x}$ and (iv) $\text{lr}_{\lambda}$ are the learning rates for the SGD optimizers of $x_i$ and $\lambda_i$, respectively; (v)~$n_\text{rec}$ is the number of reconstruction epochs; and (vi)~$m$ is the number of reconstruction candidates.

The hyperparameters $\lambda_{\text{min}}, \sigma_x, \text{lr}_{x}$ and $\text{lr}_{\lambda}$ are tuned in-training at each task $\tau$. We use three strategies: (1)~In the \emph{na{\"i}ve} strategy, we perform no hyperparameter optimization and simply run $n_\text{rec}$ reconstruction epochs with default hyperparameters based on related works from computer vision \citep{haim2022reconstructing,buzaglo2024deconstructing}. We select this strategy as our reference when we refer to our \frameworkshort, and consider the following other two strategies as ablations. (2)~In the \emph{unsupervised} strategy, we use a Bayesian search at each task to optimize the hyperparameters by minimizing the \Cref{eq:loss_full}, where each trial runs for $n_\text{rec}$ epochs. (3)~In the \emph{supervised} strategy, we use reference data of old tasks stored in a memory buffer to guide the optimization process. We then maximize the structural similarity index measure (SSIM) \citep{wang2004image} between reconstructed data and real reference data. To measure the similarity among images, we follow \citet{buzaglo2024deconstructing} and, for each real sample, find the nearest generated neighbour using the $L_2$ distance and then compute the SSIM between closest pairs. Details are in \Cref{sec:similarity_computation}. As before, we run $n_{\text{trials}}$ parameter trials at each task.

Note that hyperparameters $n_\text{rec}$ and $m$ are not tuned during training but set beforehand. We analyse their influence on \frameworkshort in sensitivity studies.

\section{Experimental Setup}\label{sec:experimental_setup}

\begin{figure*}[!htbp]
\begin{minipage}[c]{0.47\linewidth}
\includegraphics[width=\linewidth]{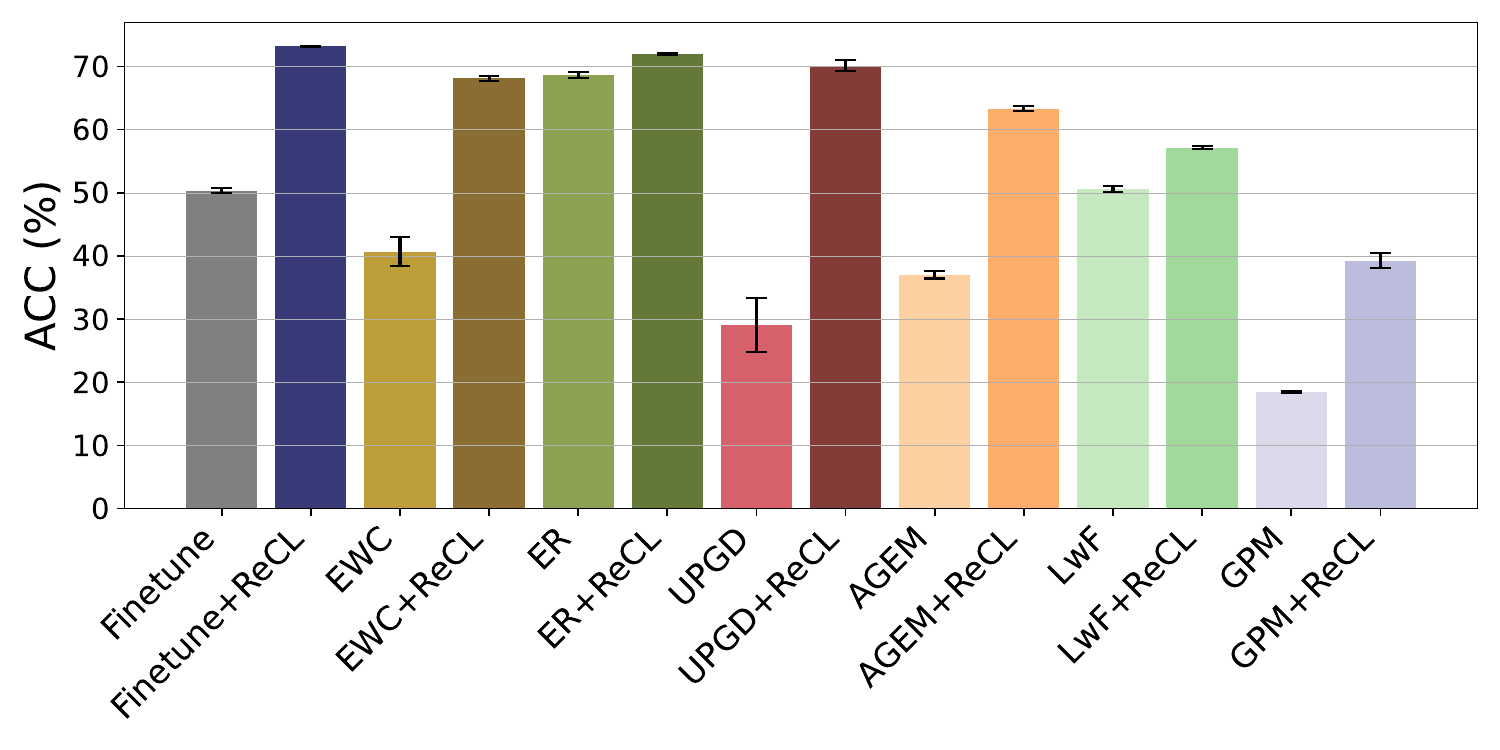}
     \caption{\textbf{ACC[$\uparrow$] for CIL, SplitMNIST:} All methods benefit from our \frameworkshort. Used standalone, \frameworkshort already is competitive to CL methods.}
     \label{fig:cil_acc_mnist}
\end{minipage}
\hfill
\begin{minipage}[r]{0.47\linewidth}
\centering
\includegraphics[width=\linewidth]{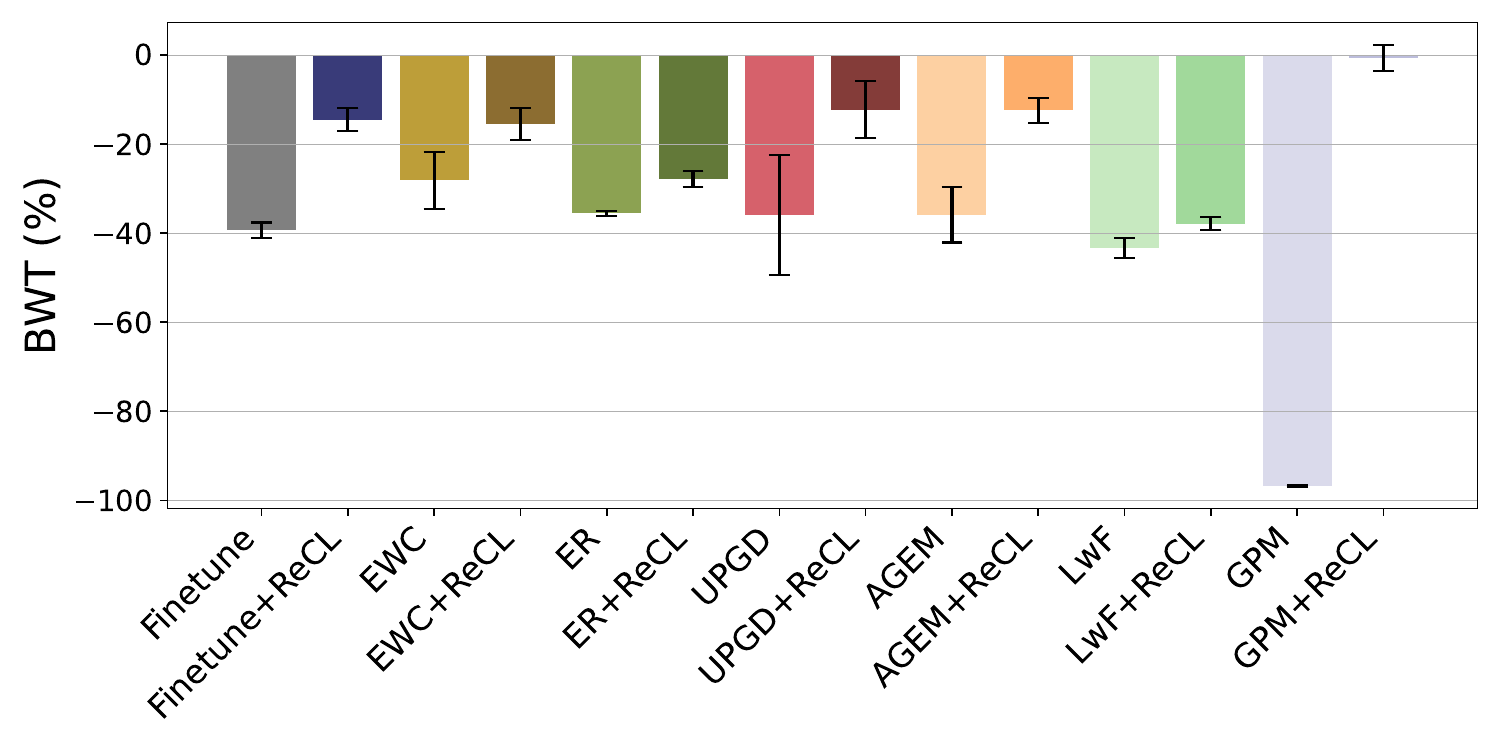}
     \caption{\textbf{BWT[$\uparrow$] for CIL, SplitMNIST:} \frameworkshort strongly reduces forgetting for all methods and is competitive when used standalone.}
     \label{fig:cil_bwt_mnist}
\end{minipage}
\end{figure*}
We adopt CL scenarios (i.e., CIL and DIL), datasets, and baselines analogous to prior literature \citep[e.g.,][]{kirkpatrick2017overcoming,serra2018overcoming,deng2021flattening,van2022three,elsayed2024addressing}. We later report ACC and BWT (see \Cref{sec:problem_definition}) over \num{5} runs with independent seeds, where, in each run, we randomly distribute the classes into tasks.

\textbf{Datasets:} We evaluate our \frameworkshort using the split variant of MNIST~\citep{lecun1998gradient}, CIFAR10~\citep{alex2009learning}, and TinyImageNet~\citep{le2015tiny,deng2021flattening}, where the classes are split into sequential tasks for CL. We refer to them as SplitMNIST, SplitCIFAR10, and SplitTinyImageNet, respectively. All datasets are well established for benchmarking in CL research (\emph{e.g.}, \citep{zenke2017continual,serra2018overcoming,deng2021flattening,rajasegaran2019random,van2022three,elsayed2024addressing}). We split the datasets into five non-overlapping tasks (two tasks with 20 classes each for SplitTinyImageNet), and randomly select $n=100$ training samples per class. We use the entire respective test set for reporting the out-of-sample performance. See \Cref{app:ds_details} for further details.

\textbf{Baselines:} We adopt a default baseline where a machine learning model is trained sequentially on each task but without reducing forgetting in any way (named \textbf{Finetune}). We further implement six state-of-the-art methods for CL: \textbf{EWC} \citep{kirkpatrick2017overcoming}, \textbf{ER} \citep{rolnick2019experience}, \textbf{UPGD} \citep{elsayed2024addressing}, \textbf{AGEM} \citep{lopez2017gradient}, \textbf{LwF} \cite{li2017learning}, and \textbf{GPM} \citep{saha2021gradient}.  

We implement the following variants of our \frameworkshort framework. (i)~We report a \frameworkshort-only baseline with a standard machine learning model, but \emph{no} other CL method for slowing down forgetting (named \textbf{Finetune+\frameworkshort}). (ii)~We further integrate the above CL baselines into our framework, which we refer to as \textbf{EWC+\frameworkshort}, \textbf{ER+\frameworkshort}, \textbf{UPGD+\frameworkshort}, \textbf{AGEM+\frameworkshort}, \textbf{LwF+\frameworkshort}, and \textbf{GPM+\frameworkshort}. Thereby, we demonstrate that our framework can be flexibly used on top of state-of-the-art CL methods. Further, any performance gain between the variant with our framework and the vanilla baselines must then be directly attributed to the fact that our framework successfully slows down forgetting.

\textbf{Implementational details:} We use SGD \citep{bottou1998online}, both for training and for data reconstruction (\Cref{eq:loss_full}), where set $n_{\text{rec}}=100$, $m=n$, and, where applicable, run $n_{\text{trials}}=100$ trials using Bayesian search with Optuna \citep{akiba2019optuna} to optimize the reconstruction hyperparameters during training (\emph{(un-)supervised strategies}). The corresponding hyperparameter grid is given in \Cref{app:recon_hparams}. To ensure a fair comparison, we extensively tuned the hyperparameters over the search grid in \Cref{sec:hparam_search_ranges}. See \Cref{app:imp_details} for more details.

\textbf{Scalability:} All experiments are conducted on a single NVIDIA A100 80GB GPU. We report average runtimes in the appendix. The overhead from adding our framework and the default na{\"i}ve strategy is relatively small. See \Cref{sec:scalability_analysis} for more details.

\section{Results}
\subsection{Scenario CIL}
\label{sec:main_cil}
\begin{table}[ht]
\centering
  \scriptsize
  \tabcolsep=0.15cm
      \caption{\textbf{Scenario CIL:} Results for training a MLP on SplitMNIST and CNN on SplitCIFAR10. Shown: average ACC[$\uparrow$] and BWT[$\uparrow$] over 5 random repetitions with varying order and init.}
      \label{tab:CIL_mlp_main}
      \centering
        \begin{tabular}{lS[table-format=2.2(2)]S[table-format=2.2(4)]S[table-format=2.2(2)]S[table-format=2.2(2)]}
\toprule
\textbf{Method} & \multicolumn{2}{c}{\textbf{SplitMNIST}} & \multicolumn{2}{c}{\textbf{SplitCIFAR10}}\\
\cmidrule(lr){2-3} \cmidrule(lr){4-5}
 & {ACC(±std)} & {BWT(±std)} & {ACC(±std)} & {BWT(±std)}\\
\midrule
Finetune & 50.34(0.42) & -39.33(2.44) & 15.87(0.14) & -71.40(0.77) \\
Finetune+\frameworkshort & 73.18(0.09) & -14.51(1.64) & 20.15(0.43) & -54.48(1.13) \\
\midrule
EWC & 40.67(2.31) & -28.18(16.37) & 16.05(0.12) & -71.13(0.94) \\
EWC+\frameworkshort & 68.13(0.41) & -15.49(2.19)  & 16.10(4.06) & -71.67(0.24) \\
\midrule
ER & 68.59(0.47) & -35.54(0.71) & 21.05(0.04) & -62.90(1.09) \\
ER+\frameworkshort & 71.94(0.11) & -27.85(2.22) & 22.46(0.27) & -57.11(0.77) \\
\midrule
UPGD & 29.10(4.26) & -35.89(12.54) & 15.94(0.13) & -71.57(0.21) \\
UPGD+\frameworkshort & 70.10(0.90) & -12.28(1.73) & 15.92(0.74) & -71.67(0.08) \\
\midrule
AGEM & 37.03(0.61) & -35.85(3.23) & 16.18(0.04) & -71.49(0.32) \\
AGEM+\frameworkshort & 63.31(0.41) & -12.43(3.95) & 16.12(4.13) & -71.28(0.50) \\
\midrule
LwF & 50.56(0.43) & -43.34(2.67) & 19.71(0.74) & -53.39(0.53) \\
LwF+\frameworkshort & 57.15(0.22) & -37.92(1.60)  & 23.59(0.38) & -40.65(3.28) \\
\midrule
GPM & 18.48(0.15) & -96.75(0.18) & 14.10(0.45) & -68.95(0.36) \\
GPM+\frameworkshort & 39.25(1.22) & -0.60(0.52) & 19.56(0.61) & -61.12(1.99) \\
\bottomrule
\end{tabular}
\end{table}
\textbf{Main results:} The main results are in \Cref{fig:cil_acc_mnist,fig:cil_bwt_mnist,tab:CIL_mlp_main} (further results are in \Cref{sec:appendix_results_cil}). We make the following observations: (\textbf{1})~Our Finetune+\frameworkshort improves over the Finetune baseline by up to \SI{45.37}{\percent} (ACC) and \SI{63.11}{\percent} (BWT). Hence, even the standalone version of our \frameworkshort framework is competitive. (\textbf{2})~Our proposed \frameworkshort is highly effective: it slows down forgetting (lower BWT) and improves classification (higher ACC) for \emph{all} baseline methods. Note that we ensured fair comparison by performing extensive hyperparameter tuning for all baselines. Hence, all performance improvements must be attributed to how our framework slows down forgetting. 

\textbf{Sensitivity study.} We conduct a sensitivity study where we (i)~vary the number of reconstruction samples (default: $m=n=100$) in the CIL scenario and where we (ii)~adopt alternative strategies for tuning the reconstruction hyperparameters. The results are in \Cref{fig:cil_ac_vs_samples} (further results are in \Cref{tab:class_sensitivity} in the appendix). We make the following observations: \textbf{(1)}~At all numbers of reconstruction samples, our framework improves over the Finetune baseline. \textbf{(2)}~We find that different tuning strategies for the reconstruction hyperparameters can lead to further performance improvements. Previously, we adopted the na{\"i}ve strategy when setting reconstruction-related hyperparameters due to the high computational overhead from the other strategies. Still, the na{\"i}ve strategy outperforms the Finetune baseline by a large margin. \textbf{(3)}~The supervised strategy outperforms the na{\"i}ve one from above. This demonstrates that reconstructed data can be optimized by supervision with reference data. \textbf{(4)}~The {unsupervised} tuning strategy for reconstruction strategy outperforms the \emph{supervised} strategy. This an interesting finding, as the unsupervised strategy does not need buffered reference data for hyperparameter tuning, which is beneficial in real-world applications.

\begin{figure}[!t]
\centering
\includegraphics[width=\linewidth]{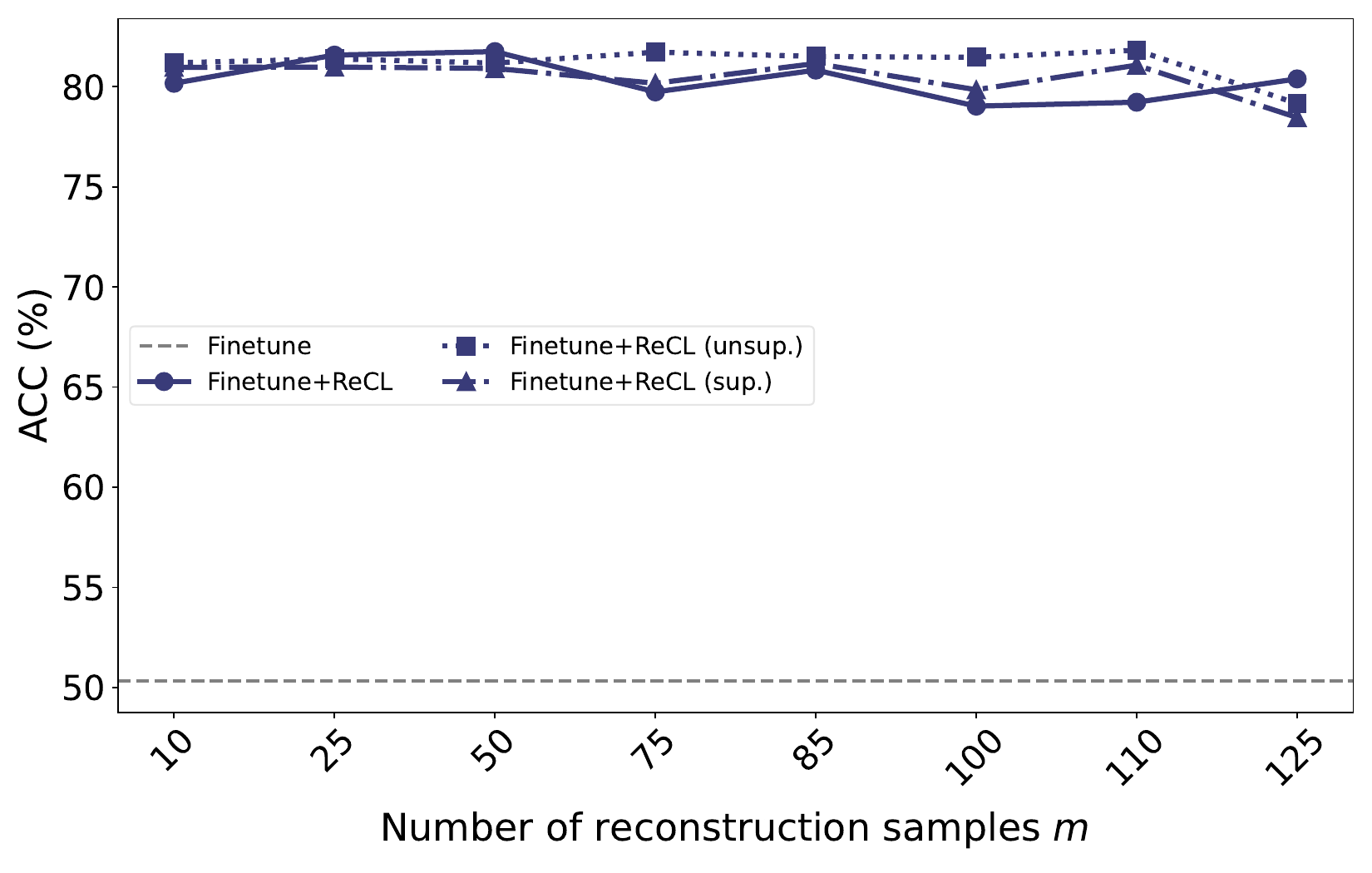}
     \caption{\textbf{Sensitivity to the number of reconstruction samples (scenario: CIL, SplitMNIST).} The performance gain increases for larger $m$, but already $m=10$ outperforms \emph{Finetune}.}
     \label{fig:cil_ac_vs_samples}
\end{figure}

\textbf{Further results:} We varied the number of reconstruction epochs from default $n_{\text{rec}}=1000$ to study the sensitivity for MLPs trained on SplitMNIST (see \Cref{sec:varying_reconstruction_epochs}). We find that we can improve ACC up to \SI{73.18}{\percent}, and already $n_{\text{rec}}=500$ reconstruction epochs lead to improvements.

$\Rightarrow$ \textbf{Takeaway}: \emph{Our \frameworkshort slows down forgetting and improves the performance of existing CL baselines.} 

\subsection{Scenario DIL}\label{sec:main_dil}

\begin{figure}[!htbp]
\begin{minipage}[c]{0.48\linewidth}
\includegraphics[width=\linewidth]{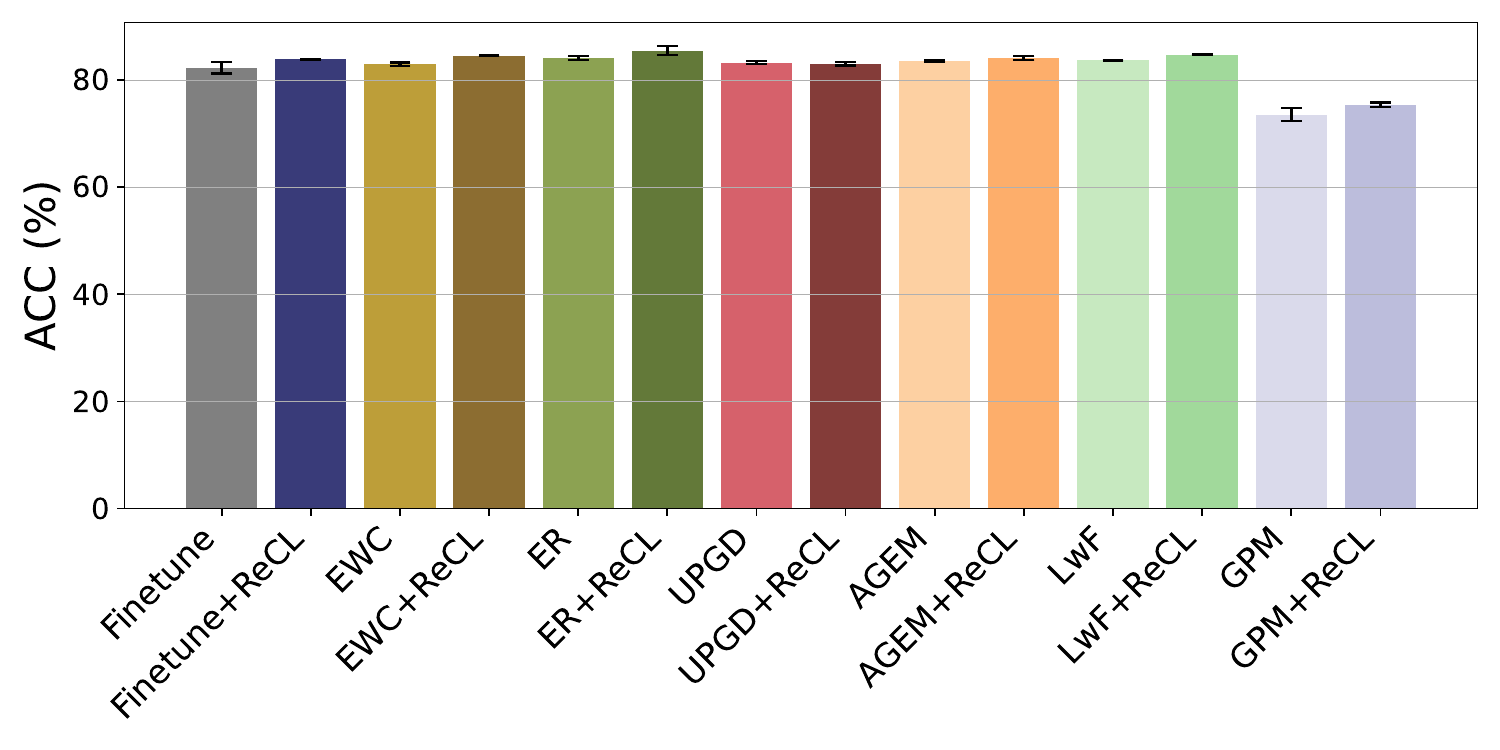}
     \caption{\textbf{ACC[$\uparrow$] for DIL, SplitMNIST:} All methods benefit from our \frameworkshort.}
     \label{fig:dil_acc_mnist}
\end{minipage}
\hfill
\begin{minipage}[r]{0.48\linewidth}
\centering
\includegraphics[width=\linewidth]{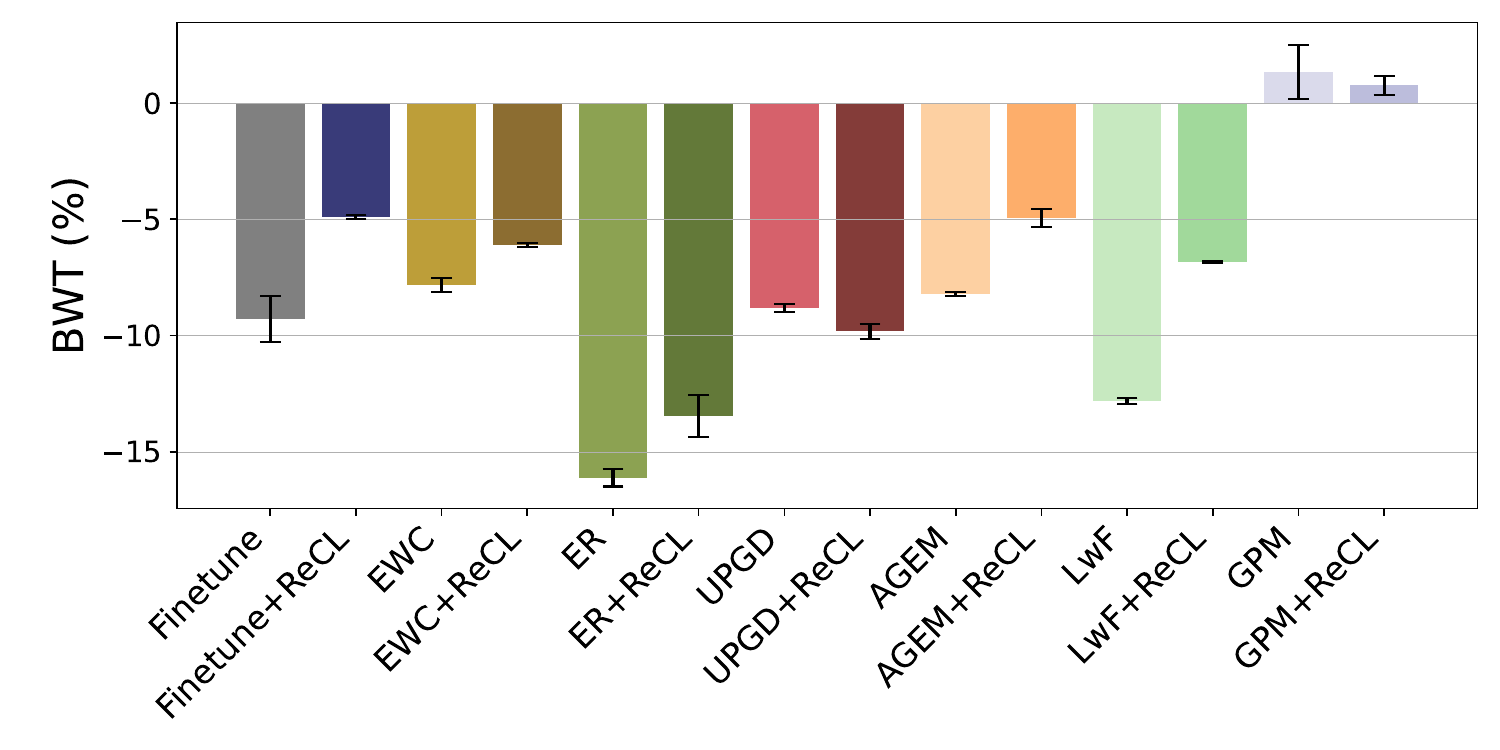}
     \caption{\textbf{BWT [$\uparrow$] for DIL, SplitMNIST:} \frameworkshort reduces forgetting for all methods.}
     \label{fig:dil_bwt_mnist}
\end{minipage}
\end{figure}
\vspace{-.15cm}
\begin{table}[h]
      \centering
      \scriptsize
      \tabcolsep=0.15cm
        \caption{\textbf{Scenario DIL:} Results for training a MLP on SplitMNIST and CNN on SplitCIFAR10. Shown: average ACC[$\uparrow$] and BWT[$\uparrow$] over 5 random repetitions with varying order and init.}
        \label{tab:DIL_mlp_main}
        \begin{tabular}{lS[table-format=2.2(2)]S[table-format=2.2(2)]S[table-format=2.2(2)]S[table-format=2.2(2)]}
\toprule
\textbf{Method} & \multicolumn{2}{c}{\textbf{SplitMNIST}} & \multicolumn{2}{c}{\textbf{SplitCIFAR10}}\\
\cmidrule(lr){2-3} \cmidrule(lr){4-5}
 & {ACC(±std)} & {BWT(±std)} & {ACC(±std)} & {BWT(±std)}\\
\midrule
Finetune & 82.23(1.09) & -9.29(1.36) & 62.59(0.69) & -10.57(2.44) \\
Finetune+\frameworkshort & 83.80(0.06) & -4.91(0.13) & 63.87(1.03) & -10.27(1.71) \\
\midrule
EWC & 82.91(0.29) & -7.82(0.31)  & 64.09(0.52) & -9.40(0.87) \\
EWC+\frameworkshort & 84.48(0.10) & -6.11(0.10) & 63.19(0.31) & -9.11(1.13) \\
\midrule
ER & 84.00(0.38) & -16.12(0.50) & 64.11(0.46) & -9.95(0.98) \\
ER+\frameworkshort & 85.44(0.91) & -13.48(1.12) & 64.08(0.61) & -8.43(0.87) \\
\midrule
UPGD & 83.19(0.23) & -8.82(0.53) & 61.39(0.51) & -13.54(1.16) \\
UPGD+\frameworkshort & 82.95(0.31) & -9.82(0.43) & 63.24(0.41) & -11.66(0.80) \\
\midrule
AGEM & 83.46(0.11) & -8.21(0.19) & 61.97(0.72) & -13.19(0.81) \\
AGEM+\frameworkshort & 84.00(0.39) & -4.95(0.96) & 62.03(1.56) & -12.54(1.47) \\
\midrule
LwF & 83.62(0.13) & -12.83(0.15)  & 65.66(0.54) & -4.17(0.74) \\
LwF+\frameworkshort & 84.67(0.04) & -6.84(0.20) & 65.59(0.45) & -1.72(0.43) \\
\midrule
GPM & 73.48(1.16) & 1.34(1.01) & 60.64(1.39) & -4.56(1.19) \\
GPM+\frameworkshort & 75.29(0.46) & 0.76(0.92) & 61.98(0.86) & -1.91(1.78) \\
\bottomrule
\end{tabular}
\end{table}
\textbf{Main results:} The results are in \Cref{fig:dil_acc_mnist,fig:dil_bwt_mnist} (with further results in \Cref{sec:appendix_results_dil}). We find: (\textbf{1})~When used standalone, our Finetune-\frameworkshort slows down forgetting in comparison to the Finetune baseline by up to \SI{1.92}{\percent} (ACC) and \SI{47.15}{\percent} (BWT). (\textbf{2})~Our proposed \frameworkshort is again highly effective: it slows down forgetting (higher BWT) and improves classification (higher ACC) for existing CL methods.

\textbf{Further results:} \textbf{(1)}~The results for the sensitivity study are in \Cref{sec:appendix_dil_sensitivity}. We find that the performance is high across all values, with the default \num{100} samples performing best. The na{\"i}ve reconstruction strategy performs similar to the optimization strategies while adding considerably less computational overhead. \textbf{(2)}~We further varied the number of reconstruction epochs (see \Cref{tab:dil_varying_reconstruction_epochs}) and find that \num{1000} reconstruction epochs performs best, with no improvements from ablated settings. Overall, our \frameworkshort framework leads to performance gains.

\subsection{TinyImageNet}\label{sec:tinyimagenet}
\begin{figure}[!hbtp]
\includegraphics[width=\linewidth]{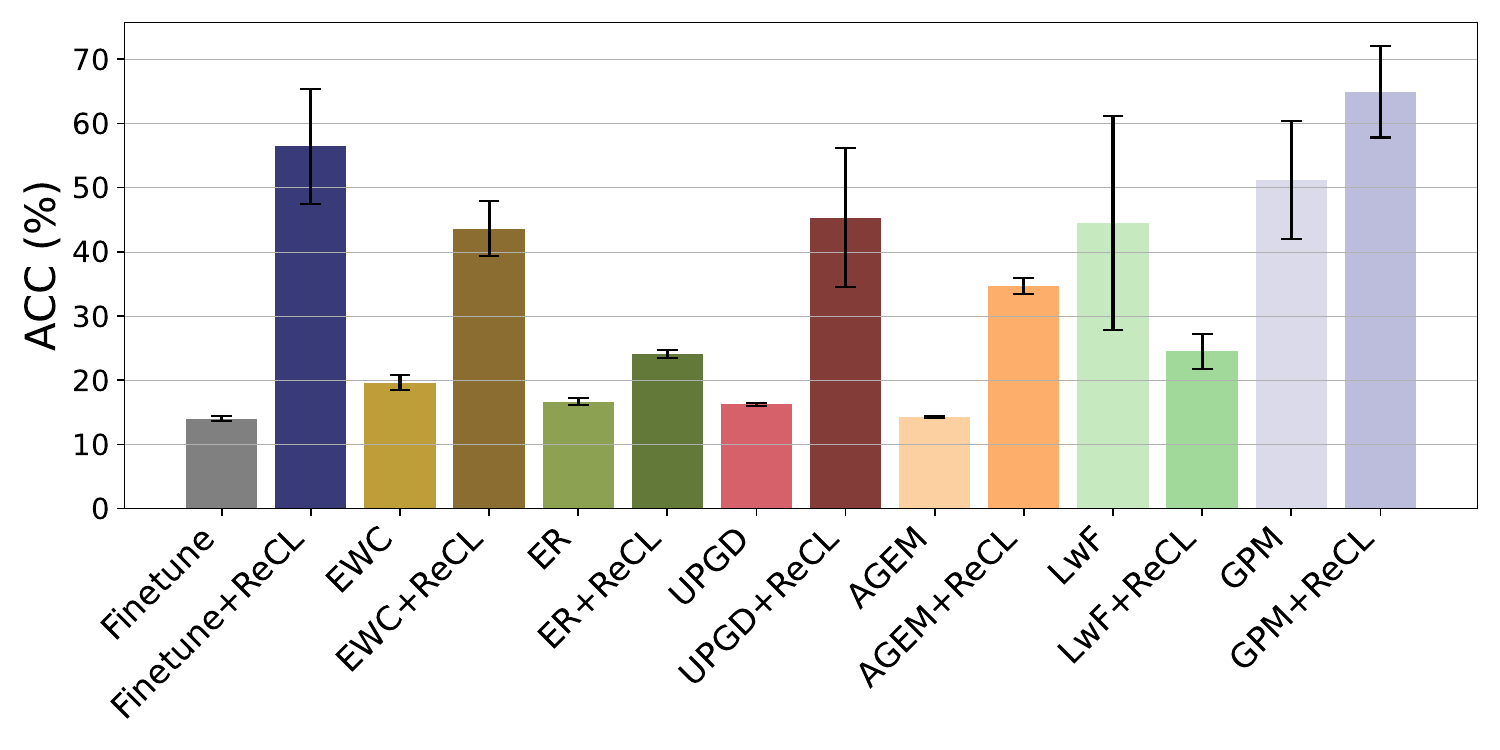}
     \caption{\textbf{ACC [$\uparrow$] for DIL, TinyImageNet} All methods benefit from our \frameworkshort. Used standalone, \frameworkshort already is competitive to CL methods.}
     \label{fig:tinyimagenet_domain_main}
\end{figure}
\textbf{Main results}: The results for SplitTinyImageNet are in \Cref{fig:tinyimagenet_domain_main} (further results in \Cref{app:tinyimagenet}) and align with our previous findings. Notably, we find: \textbf{(1)} Used standalone, \frameworkshort strongly improves the performance by \SI{42}{\percent}p. \textbf{(2)} \frameworkshort also improves existing CL methods in both the CIL and DIL scenarios.

$\Rightarrow$ \textbf{Takeaway}: \emph{Our \frameworkshort improves the performance of existing CL baselines on real-world data.} 

\subsection{Further Insights}

\textbf{Can \frameworkshort be used for online CL?}
\begin{figure}
    \centering
    \includegraphics[width=\linewidth]{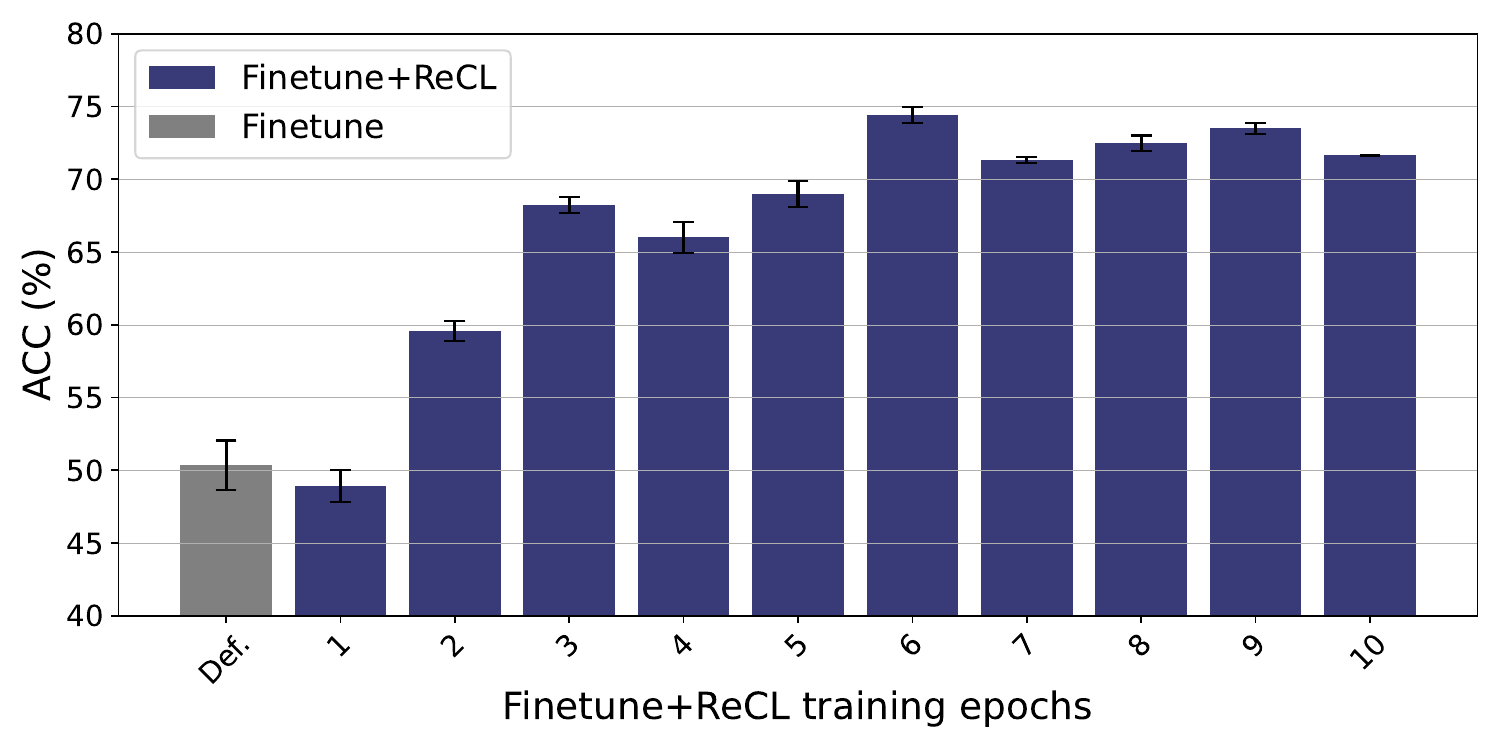}
    \caption{Ablation on train epochs for \frameworkshort on \textbf{MNIST} in scenario \textbf{CIL}. Def. = default finetune performance without \frameworkshort.}\label{fig:abl_mnist_ft_epochs}
\end{figure}
For this, we evaluate \frameworkshort in the challenging CIL scenario. Recall that in CIL, each task introduces novel classes. Additionally, in an online setup, each class' samples are seen a limited number of times only, further increasing the difficulty. To simulate this, we deliberately vary the epochs from 1 to 10, thereby varying the strictness of the online CL. The results in \Cref{fig:abl_mnist_ft_epochs} show that strict online CL is not possible, but, already after two passes over the data, our \frameworkshort improves the performance. Generally, more training epochs (i.e., more passes over the training data) increase the performance of \frameworkshort.

\textbf{Can \frameworkshort slow down forgetting even for non-homogenous networks?} For this, we use \frameworkshort and now train AlexNet \citep{krizhevsky2012imagenet} and ResNet~\citep{he2016deep} in the DIL scenario. Both architectures are \emph{not} homogenous neural networks because they use modules that introduce ``jumps'' into the propagation of hidden activations. AlexNet uses max-pooling, dropout~\citep{srivastava2014dropout} and bias vectors, and ResNet uses residual connections and batch normalization~\citep{ioffe2015batch}. Both architectures thus violate the theoretical foundation of our CL framework. Nonetheless, we find that our \frameworkshort is effective: it improves ACC by \SI{4.77}{\percent} (AlexNet) and $\sim$\SI{3}{\percent} (ResNet, \Cref{tab:alexnet_results}; details in \Cref{sec:alexnet_results}). $\Rightarrow$~\textbf{\emph{Takeaway:}} \emph{Our \frameworkshort is also effective for non-homogenous network architectures where it can slow down forgetting successfully}.

\begin{table}[h!]
\footnotesize
\tabcolsep=0.11cm
\centering
\caption{\textbf{Scenario DIL:} Results for using our framework with AlexNet \citep{krizhevsky2012imagenet} and ResNet18 \citep{he2016deep}, two \emph{non}-homogenous neural networks violating the theoretical background of our framework. We nonetheless see improvements from using our \frameworkshort.}
\label{tab:alexnet_results}
\footnotesize
\begin{tabular}{lSSS}
\toprule
\textbf{Method} & \multicolumn{2}{c}{\textbf{SplitCIFAR10}}\\
\cmidrule(lr){2-3}
{} & {ACC} & {BWT} \\
\midrule
AlexNet & 56.47 & -31.63\\
AlexNet+\frameworkshort & \textbf{59.23} & -28.25\\
\midrule
ResNet & 53.53 & -28.17 &\\
ResNet+\frameworkshort & \textbf{55.16} & -29.41 \\
\bottomrule
\end{tabular}
\end{table}

\begin{figure}
\begin{subfigure}{0.49\linewidth}\includegraphics[width=\linewidth]{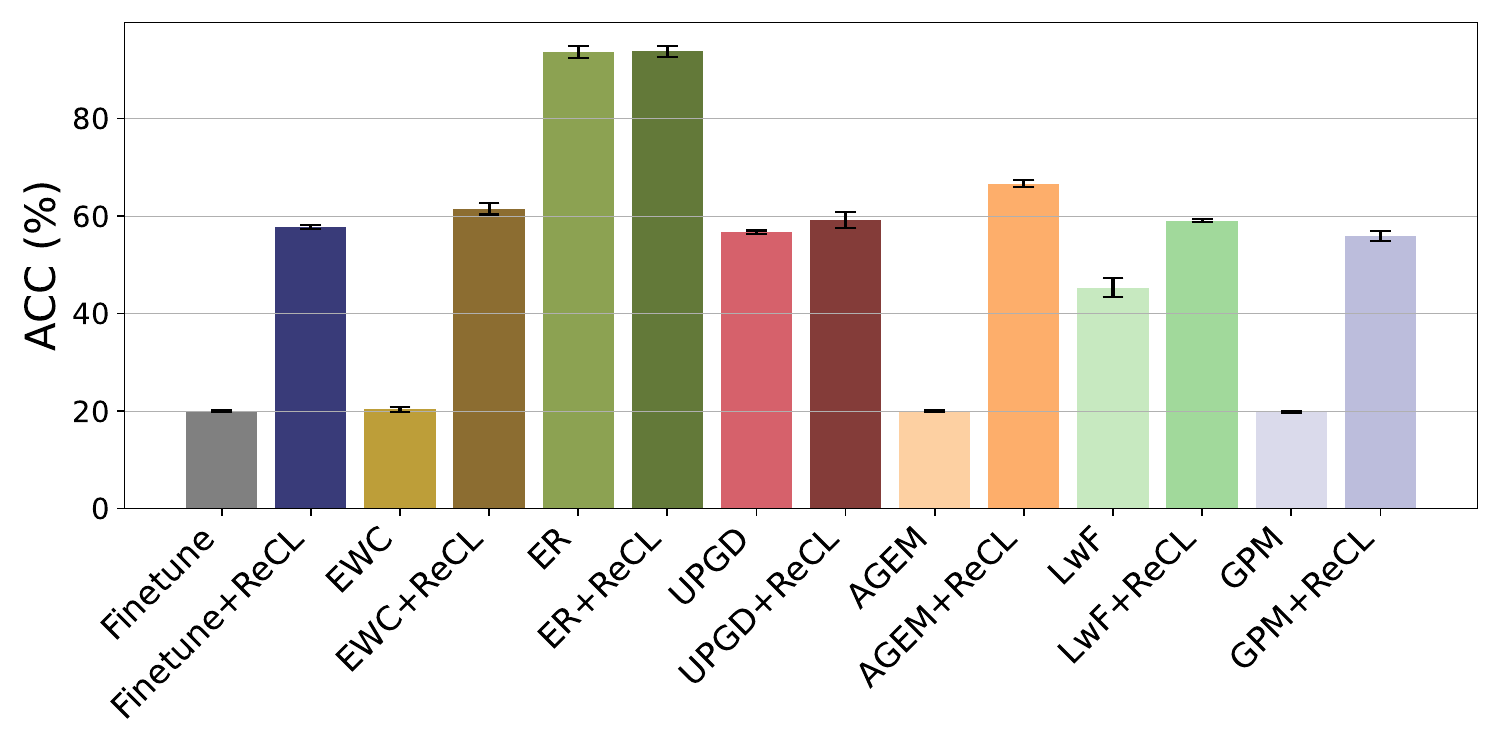}
    \caption{\emph{full} \textbf{SplitMNIST}}
\end{subfigure}
\begin{subfigure}{0.49\linewidth}
\includegraphics[width=\linewidth]{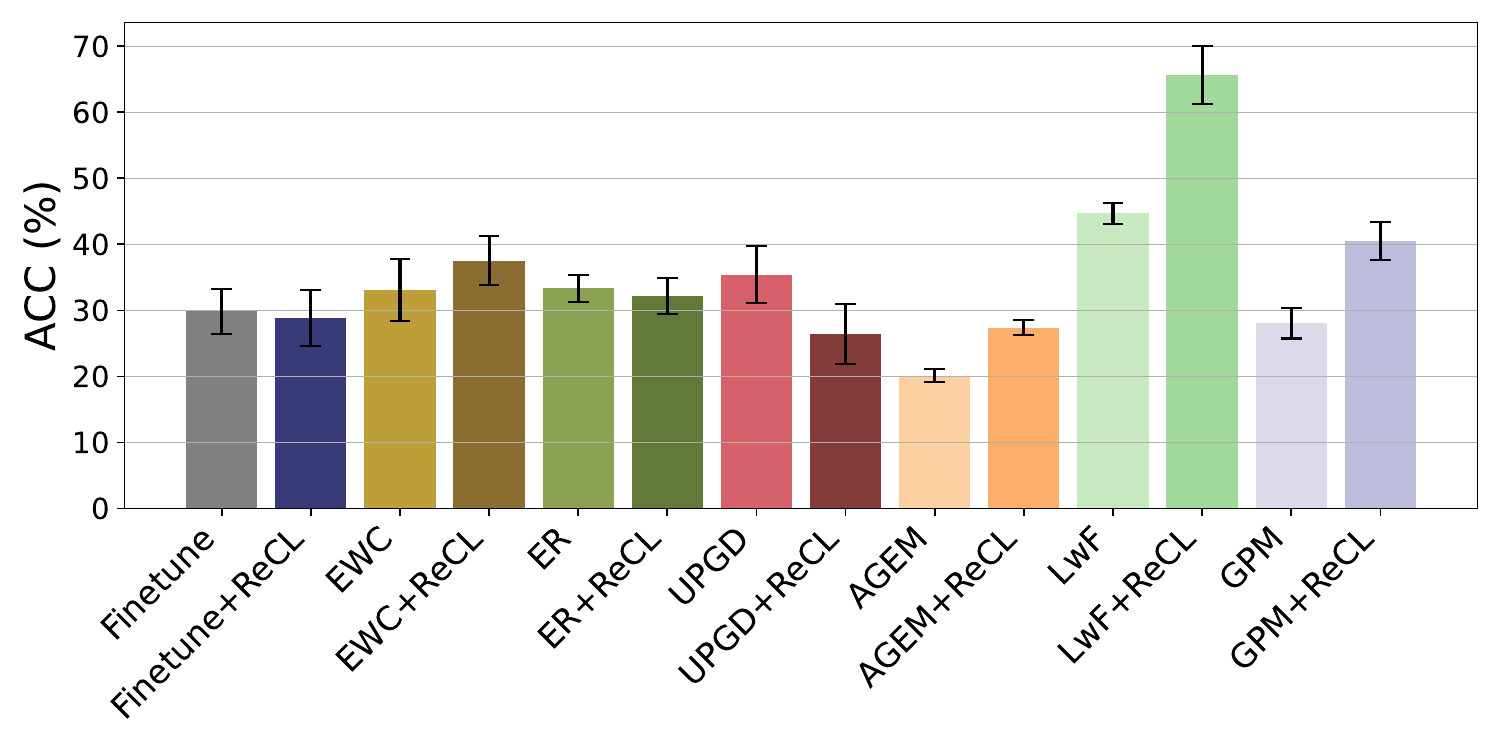}    \caption{\emph{full} \textbf{SplitTinyImageNet}}
\end{subfigure}
\caption{ACC for \frameworkshort on \emph{full} SplitMNIST and SplitTinyImageNet datasets.}
\label{fig:full_ds_results}
\end{figure}

\textbf{Extended results for the full datasets.}
We now provide extended results for full datasets. For this, we select SplitMNIST and SplitTinyImageNet and use the entire training data. The results are visualized in \Cref{fig:full_ds_results}. Our observations align with our previous findings. We again find that our framework is highly effective, regardless of the size of the dataset. \frameworkshort can improve the performance of methods that are known to underperform (e,g, EWC in CIL \citep{van2019three}) more than \SI{40}{\percent}p. At the same time, \frameworkshort reduces the BWT by up to \SI{122}{\percent}, sometimes even leading to \emph{retrospective} performance improvements on old tasks. Detailed results  are in \Cref{sec:scaling_recl}. $\Rightarrow$~\textbf{\emph{Takeaway:}} \emph{Our \frameworkshort can successfully slow down forgetting for large datasets.}

\section{Conclusion}
In this paper, we propose a novel framework for continual learning, termed \frameworkshort (\frameworklong), which slows down forgetting by using the neural network parameters as a built-in memory buffer. Different from existing approaches, our framework does not require separate generative networks or memory buffers. Instead, our \frameworkshort framework offers a novel paradigm for dealing with forgetting by leveraging the implicit bias of convergence to margin maximization points. Through extensive experiments, we show that \frameworkshort is compatible with (i)~existing CL methods, (ii)~different datasets, (iii)~different dataset sizes, and (iv)~different network architectures. Further, we also show the applicability of \frameworkshort in challenging continual learning scenarios, namely class- and domain-incremental learning. Across all experiments, we observe consistent performance gains through our framework. Lastly, we expect our framework to be of practical value for real-world applications, where computational (e.g., sheer data size) or regulatory reasons only allow carrying over a limited amount of data from old tasks.

{
    \small
    \bibliographystyle{ieeenat_fullname}
    \bibliography{main}
}

\clearpage
\appendix
\setcounter{page}{1}
\maketitlesupplementary
\renewcommand{\thesection}{\Alph{section}}

\section{Extended Literature Review}\label{app:extendend_lit}
\textbf{Continual Learning:} CL learning methods are often categorized into three categories, namely \emph{memory-based}, \emph{architecture-based}, and \emph{regularization-based} \citep{deng2021flattening,saha2021gradient,wang2024comprehensive}.

\textbf{Memory-based methods.} Methods using memory-based techniques either maintain a (small) amount of old task's data (or separate datasets) in a memory buffer or train generative models for old tasks. OrDisCo \citep{wang2021ordisco} uses a setup consisting of a generator, discriminator, and classifier to pseudo-label unlabelled data from a separate dataset and train these components together. iCarl \citep{rebuffi2017icarl} splits the feature and the classification parts of the network, and stores samples that are closest to the feature mean of a class. Variants of the gradient projection memory (GPM) approach \citep{deng2021flattening,saha2021gradient} store vectors describing the core gradient space in memory and guide update to be orthogonal to this space. In deep generative replay, \citet{shin2017continual} use a dual-network architecture consisting of an additional generative part that can be sampled for old tasks' data.

\textbf{Architecture-based methods.} Methods based on architectural  modifications allocate parts of a (growing) model to specific tasks. Progressive Neural Networks (PNN) \citep{rusu2016progressive} adds a separate network branch for new tasks and utilizes previous knowledge via lateral connections to previous tasks' frozen branches, avoiding forgetting at the cost of growing network size. SupSup by \citet{wortsman2020supermasks} uses trained task-specific binary masks overlaid over a fixed random neural network to allocate parameters. The Winning Subnetworks (WSN) \citep{kang2022forget} approach follows a similar route, but allows training of the underlying neural network and restricts mask sizes. Finally, PackNet \citep{mallya2018packnet} prunes a task's mask after training and then Finetunes the underlying weights while keeping parameters of previous tasks intact. APD \citep{yoon2020scalable} splits parameters into task-shared and task-specific parameters. They enforce sparsity on the task-specific parameters and retroactively update them to respond to changes in task-shared parameters.

\textbf{Regularization-based methods.} Methods in the regularization category use additional regularization terms in the loss function to restrict updates to important parameters. Examples here are elastic weight consolidation (EWC) \citep{kirkpatrick2017overcoming}, which uses the Fisher information matrix to assess parameter importance, and Learning without Forgetting (LwF) \citep{li2017learning}, which records the responses of the network to a new task's data prior to training and tries to keep recorded and newly learned responses similar. \citet{lee2019overcoming} employ multiple distillation losses to distill the previous task's model into a model combining the old and current tasks.

\textbf{Additional categories}. Beyond these three widely used categories, some taxonomies use five categories, adding the (i)~\emph{optimization} and (ii)~\emph{representation} approaches \citep{wang2024comprehensive}.
Methods belonging to the \textbf{(i)~optimization} category manipulate the optimization process. The already introduced GPM approaches \citep{deng2021flattening,saha2021gradient} also fall into this category, as they modify gradient updates. Another method is StableSGD \citep{mirzadeh2020understanding}, which optimizes hyperparameters such as the learning rate in order to find flat minima which potentially ``cover'' all tasks. Methods belonging to the \textbf{(ii)~representation} category focus on utilizing the representations learned during deep learning. Self-supervised learning and pre-training fall into this category, as they can provide a good initialization of neural networks for sequential downstream tasks \citep{madaan2022representational,shi2022mimicking,wu2022class}. Other methods explore continual pre-training, building on the fact that pre-training data might be collected incrementally \citep{han2021econet,cossu2024continual}.

\textbf{Implicit Bias and Data Reconstruction:} Neural networks can generalize to unseen (test) data. This phenomenon is described as an \emph{implicit bias} of neural network training, and it has been studied extensively \citep{gunasekar2018characterizing,amid2020reparameterizing,blanc2020implicit,moroshko2020implicit,azulay2021implicit,damian2021label,li2021happens,wang2021implicit,nacson2022implicit,nacson2023implicit}. For a certain types of neural networks, described in \Cref{app:homogenous_definition}, the implicit bias causes network weights to converge to margin maximization points \citet{lyu2019gradient,ji2020directional}. \citet{haim2022reconstructing} show that under this condition, the entire training data can be recovered from pretrained binary classifiers. \citep{loo2023understanding} utilized this for networks trained under the neural tangent kernel regime, and \citet{buzaglo2024deconstructing} extend the underlying theory to multi-class networks and general loss functions.

\newpage
\section{Background on Dataset Reconstruction}\label{app:ds_recon_background}
In this section, we detail the background of the dataset reconstruction process, based on preliminary works \citep{lyu2019gradient,ji2020directional,haim2022reconstructing,buzaglo2024deconstructing}. We start with the definition of \emph{homogenous neural networks} and then give details on dataset reconstruction.

\subsection{Homogenous Neural Networks}\label{app:homogenous_definition}
Homogenous networks are networks whose architectures satisfy the following condition \citep{lyu2019gradient}:
\begin{equation}
    \forall c > 0 : \Phi(c\theta; x) = c^L\Phi(\theta; x) \text{ for all } \theta \text{ and },
\end{equation}
if there is a $L>0$. That is, scaling parameters $\theta$ is equal to scaling the network's output $\Phi(\theta; x)$. Fully connected and convolutional neural networks with ReLu and without skip-connections and bias vectors satisfy this condition. However, popular models ResNet \citep{he2016deep}, ViT \citep{dosovitskiy2020image}, or CLIP \citep{radford2021learning} models do not satisfy this condition since they contain modules (e.g., skip connections) that restrict the propagation of scaled weights' intermediate outputs through the network. \citet{oz2024reconstructing} show that in such cases, compute- and time-intensive workarounds relying on model inversion attacks \citep{tumanyan2022splicing} and diffusion models \citep{sohl2015deep} could be applicable.

\subsection{Dataset Reconstruction}\label{sec:app_ds_recon}
\textbf{Maximum margin points.} To reconstruct the training data of past task, we build upon the implicit bias of gradient descent training \citep{vardi2023implicit}. For a homogenous neural network $\Phi(\theta,\cdot)$ trained under the gradient flow regime, \citet{lyu2019gradient} show that the weights converge to the following maximum margin point
\begin{multline}
    \min_{\theta} \frac{1}{2} \|\theta\|^2 \quad \text{s.t.} \quad \Phi_{y_i}(\theta; x_i) - \Phi_j(\theta; x_i) \geq 1 \quad \\ \forall i \in [n], \forall j \in [C] \setminus \{y_i\},
\end{multline}
where $\Phi_j (\theta; x) \in \mathbb{R}$ is the output of the $j$-th neuron in the output layer for an input $x_i$ and $[C]$ is the set of all classes (i.e., the label space). At such a point, the difference between the neuron activation of a sample's true class, $\Phi_{y_i}(\theta; x_i)$, and all other neurons, $\Phi_j(\theta; x_i)$, is maximized.

\textbf{Conditions.} This convergence point is characterized by, among others, the existence of $\lambda_1, \dots, \lambda_m \in \mathbb{R}$ and the following \emph{stationarity} and \emph{feasibility} conditions~\citep{buzaglo2024deconstructing}:
 \begin{align}
    &\theta - \sum_{i=1}^m \sum_{j\ne{y_i}}^c\lambda_{i,j}  \nabla_{\theta}  ( \Phi_{y_i}(\theta; x_i) - \Phi_{j}(\theta; x_i) ) = 0\label{eq:mult_stationary}\\
    &\forall i \in [n], \forall j \in [C] \setminus \{y_i\}: \;\;  \lambda_{i,j} \geq 0\label{eq:mult_dual feas}
\end{align}
These equations state that a (trained) neural network's parameters $\theta$ can be approximated by linearly combining the $\lambda$-scaled derivatives on data points $x_i$. Keeping $\theta$ fixed, we utilize this to optimize samples $x_i$ and coefficients $\lambda_{i,j}$ using gradient descent. Note the following two things: (i)~ The corresponding labels $y_i$ are chosen to match the number of output neurons of the pre-trained neural network\footnote{\emph{E.g.}, if the network has three output neurons, then the class labels are $[1, 2, 3]$.}. We evenly split the number of classes among the samples to be reconstructed, but uneven splits are equally possible. (ii)~The number of samples to reconstruct, $m$, generally is unknown. In this work, we consider it appropriate to maintain count of the number of samples of previous tasks. This could by achieved by, \emph{e.g.}, a small buffer which simply stores these counts. Determining the ``correct'' $m=n$ without prior knowledge is an open point and beyond the scope of this paper. As a starting point, we could imagine searching over a set of $m$ and monitoring the behaviour of the reconstruction loss across trials.

\textbf{Reconstruction loss:} We reconstruct the old training data by optimizing randomly initialized data points $x_i$ to closely resemble the old task's data via the following combined reconstruction loss $L_{\text{full}}$ \citep{buzaglo2024deconstructing}:
\begin{multline}
    L_{\text{rec}}(x_1, \ldots, x_m, \lambda_1, \ldots, \lambda_m) =\\ \left\| \theta - \sum_{i=1}^{m} \lambda_i \nabla_{\theta} \Phi(\theta; x_i ) \right\|_2^2 + 
    {L_{\lambda}} + L_\text{prior},\label{app:loss rec}
\end{multline}
where $ \left\|\cdot\right\|_2^2$ denotes the squared $\text{L}_2$ norm (i.e., squared Euclidean distance). 
\textbf{Lambda loss:} The lambda loss $L_{\lambda}$ constrains the coefficients of the linear combination via 
\begin{equation}
    L_{\lambda} = \sum_{i=1}^{m} \text{max} \left\{ -\lambda, -\lambda_{\text{min}}\right\},
\end{equation}
where $\lambda_{\text{min}}$ is a hyperparameter.

\textbf{Prior loss:} The prior loss $L_{\text{prior}}$ constrains the value range of the (reconstructed) images to lie within normalized range $[-1, 1]$.

\textbf{Optimizing $x_i$ and $\lambda_i$:} We optimize the randomly initialized samples and coefficients via stochastic gradient descent. To backpropagate the gradients through $\Phi$, we follow \citet{haim2022reconstructing} and replace ReLU activations with a smooth ReLU function in the backwards pass only\footnote{In practice, we deepcopy $\Phi$ and on the copy replace ReLU activations with a smooth counterpart.}

\subsection{Hyperparameters of the Reconstruction Process}\label{app:recon_hparams}
The reconstruction process has several hyperparameters which influence reconstruction quality. First, the reconstructed samples $x_i$ are initialized from a Gaussian distribution scaled by $\sigma_x$. In our in-training hyperparameter searches, we draw this parameter from a logarithmic uniform distribution between $[10^{-5},1]$. The $x_i$ are optimized gradient descent, with a learning rate log-uniform in $[10^{-5},1]$. For the separately-optimized coefficients $\lambda_i$, the learning rate is drawn from the same range. The $\lambda_{\text{min}}$ from \Cref{eq:loss_rec} is drawn log-uniform from $[0.01,0.5]$. To optimize the hyperparamaters in the \emph{(un-)supervised} strategies detailed in \Cref{sec:hyperparameter_tuning_strategies}, we run $n_{\text{trials}}=100$ search trials using Bayesian search via Optuna \citep{akiba2019optuna}. As default, we use $n_{\text{rec}}=1000$ reconstruction epochs at each trial.

\section{Dataset Details}\label{app:ds_details}
In this section, we give details about the three datasets used. All datasets are commonly used in previous works, \citep[e.g.,][]{kirkpatrick2017overcoming,serra2018overcoming,deng2021flattening,van2022three,elsayed2024addressing}. Below, we provide information about each of them.

\textbf{MNIST:} The MNIST \citep{lecun1998gradient} dataset contains \num{60000} black-and-white \num{28}x\num{28} training images evenly split across \num{10} classes. An additional \num{10000} samples are used for testing. The images show hand-written digits.

\textbf{CIFAR10:} The CIFAR10 \citep{alex2009learning} dataset contains \num{60000} \num{32}x\num{32} colour images evenly split across \num{10} classes. \num{50000} samples are reserved for training and \num{10000} samples for testing. This dataset depicts animals and vehicles.

\textbf{TinyImagenet}: The TinyImageNet \citep{le2015tiny} dataset contains \num{100 000} \num{64}x\num{64} colour images from \num{200} classes. Note that test labels are unavailable. We thus follow \citep{deng2021flattening} to create the training and testing splits. Further, we use a 40-class, two-task subset, with each task having 20 classes.

\textbf{Pre-processing:} We normalize the data by removing the training-data mean from each dataset.

\textbf{Dataset Splitting:} We split the datasets into $n_{\text{tasks}}=5$ (value depending on the experiment) tasks by splitting the class labels without overlap. In the domain incremental learning (DIL) scenario, we dynamically re-label the classes to begin at zero. This is done to compute the cross-entropy loss of, e.g. classes $[5,6,7,8]$, on a model with an output layer of size four (which would expect labels $[1, 2, 3, 4]$).

\newpage
\section{Implementation details}\label{app:imp_details}
In this section, we provide details about our implementation and the CL methods that we combine \frameworkshort with.

\subsection{Models}
We train \num{4}-layer multi-layer perceptron (MLP) and convolutional neural network (CNN) models. The MLP architecture is structured as \emph{D}-\num{1000}-\num{1000}-\emph{C}, where \emph{D} is the (flattened) image input and \emph{C} is the number of classes. The CNN architecture roughly follows \citep{deng2021flattening}. It uses two convolution layers with \num{160} kernels of size \num{3}, leading to CONV(\emph{k}=\num{3},\emph{ch}=\num{160})-CONV(\emph{k}=\num{3},\emph{ch}=\num{160})-\num{640}-\num{320}-\emph{C}. Following the homogenous network constraint from \Cref{sec:related_work} we use no bias vectors except for the first layer, whose weights are scaled by $10^{-4}$. Note that for both network types, the number of neurons \emph{C} in the final classification layer might be expanded over the course of incremental training.

\subsection{Baseline Details}
We implemented our \frameworklong{} (\frameworkshort) framework and the CL baselines in PyTorch. For this, we carefully considered available reference (UPGD \cite{elsayed2024addressing}, \href{https://github.com/mohmdelsayed/upgd}{https://github.com/mohmdelsayed/upgd}) and community implementations (EWC \citep{kirkpatrick2017overcoming}, \href{https://github.com/moskomule/ewc.pytorch/tree/master}{https://github.com/moskomule/ewc.pytorch}). We obtained written permission by the respective code maintainers. The same applies to the base code for the dataset reconstruction, which we derived from \href{https://github.com/gonbuzaglo/decoreco}{https://github.com/gonbuzaglo/decoreco}.

\textbf{Elastic weight consolidation:}: Elastic weight consolidation (EWC) \citep{kirkpatrick2017overcoming} is a regularization-based strategy that uses the Fisher Information Matrix to measure the importance of neural network parameters for a task. Based on the idea that the solution for a (new) task can be found in the neighbourhood of the current weights, it constrains parameters to stay close to their old value by adding the difference as a quadratic penalty. We refer to \citet{kirkpatrick2017overcoming} for details.

In both CL scenarios, we compute parameter importance after training on task $\tau$ from $\mathcal{D}_{\tau}^{tr}$. In the CIL scenario, which expands the output layer, we only constrain those values that existed before the expansion (i.e., the penalty for changing newly added parameters is zero).

\textbf{Experience replay:} Experience replay (ER) \citep{rolnick2019experience,chaudhry2019tiny} is a memory-based strategy that stores a small subset of the old tasks' training data in a memory buffer. It then replays this buffered data in training on task $\tau$. ER has originally been proposed in reinforcement learning to help an agent learn from previous experience.

In all experiments, we set a default buffer size of \SI{10}{\percent} by selecting a random subset of a task's training data for storage in memory. We do not employ any optimization strategies (such as choosing the hardest or most informative data points) when adding to or drawing from the memory buffer.

\textbf{Utility-perturbed gradient descent:} Utility-perturbed gradient descent (UPGD) \citep{elsayed2024addressing} is a regularization-based and optimization-based strategy based on the idea of pruning (un-)important parameters. It directly modifies a parameter's gradient as follows. (i)~First, it computes the performance difference between the network as-is and with the parameter set to zero. However, this is computationally prohibitive as it would require a separate forward pass for each parameter. UPGD therefore approximates the true utility of a parameter using Taylor approximation. (ii)~Second, it scales per-parameter gradients based on their utility. For more useful (=important) parameters, the gradient update size is reduce, while for non-useful paramters it is not modified. Crucially, UPGD adds Gaussian noise to the gradient. We refer to \citet{elsayed2024addressing} for details.

\textbf{Average-gradient episodic memory:} AGEM \citep{chaudhry2018efficient} is a replay- and optimization-based strategy. It stores a small number of samples (called \emph{experiences}) in memory. During training on the current task, AGEM uses the stored old samples as an additional loss component, allowing that the average loss over all previous tasks does not increase.

\textbf{Learning without Forgetting:} Learning without Forgetting (LwF) \citep{li2017learning} is an optimization-based strategy that, at the beginning of each task, stores reference outputs of the model for the current task's data. During training, the strategy penalizes changes to the network that cause the output deviate too stronly from the reference outputs. It thus searches for parameters that encode the current information and simultaneously avoid catastrophic inference.

\textbf{Gradient projection memory:} Gradient projection memory (GPM) \citep{saha2021gradient} is a replay- and optimization-based strategy. GPM computes the bases of the weight space spanned by the model's parameters, and updates the set of bases at each new task. Gradients are then projected to be orthogonal to the current space. 

\textbf{Combination with \frameworkshort:} We combine \frameworkshort with the above listed CL baseline methods by adding it on top. For this, no modifications are necessary as the data reconstruction is independent. This highlights the flexibility of our framework.

\begin{table*}[tb]
    \caption{Hyperparameter search space. The search ranges for the EWC and UPGD parameters are from \citep{elsayed2024addressing}. The hyperparameter search is conducted separately for each CL scenario, architecture, and dataset over 100 trials. ``{For}'' denotes the methods for which we optimized the listed hyperparameters.}
    \label{tab:table_search_space}
    \centering
    \footnotesize
    \begin{tabular}{lll}
    \toprule
    Parameter     &  Search Space & For\\
    \midrule
    Learning rate & 0.0001 to 0.1 (log uniform) & All\\
    Train epochs & 10 to 100 & All\\
    \midrule
    EWC-$\lambda$ & \{10, 100, 500, 1000, 2500, 5000, 10000\} & EWC\\
    \midrule
    Replay fine-tuning epochs & 1 to 10 & ER\\
    \midrule
    Patterns per experience & \{1, 10, 25, 50\} & AGEM\\
    \midrule
    GPM thresholds & \{0.9, 0.99, 0.999\} & GPM\\
    \midrule
    LwF-$\lambda$ & \{0.1, 0.5, 1, 2, 4, 8, 10\} & LwF\\
    LwF temperature & \{0.5, 1, 2, 4\} & LwF\\
    \midrule
    UPGD-$\beta$ & 0.9 to 0.9999 & UPGD\\
    UPGD-$\sigma$ & 0.0001 to 0.01 & UPGD\\
    \midrule
    Extraction epochs & \{100, 500, 1000\} & ReCL\\
    Extraction learning rate & 0.0001 to 1.0 (log uniform) & ReCL\\
    Extraction lambda learning rate & 0.0001 to 1.0 (log uniform) & ReCL\\
    Extraction min lambda & 0.01 to 0.5 (log uniform) & ReCL\\
    Extraction model ReLU alpha & 10 to 500 & ReCL\\
    Extraction init scale & 0.000001 to 0.1 (log uniform) & ReCL\\
    Hyperparameter num trials & \{10, 25, 50, 100, 150\} & ReCL\\
    \bottomrule
    \end{tabular}
\end{table*}

\subsection{Hyperparameter search ranges}\label{sec:hparam_search_ranges}
We optimized the hyperparameters of each CL method (EWC, UGPD, ER) and our baselines separately over 100 trials on a held-out validation dataset (a subset of the training data). Optimized hyperparameters and their search ranges are given in \Cref{tab:table_search_space}.

\subsection{Training Details}\label{sec:training_details} 
All networks are trained with stochastic gradient descent (SGD) \citep{bottou1998online} with a batch size of \num{64}. We use the cross-entropy loss based on logits (i.e., no activation on $\phi_h$, following \Cref{app:homogenous_definition}).We repeat each experiment \num{5} times, each time with a different seed.

In preliminary experiments, we followed previous works \citep{haim2022reconstructing,buzaglo2024deconstructing} and used full-batch gradient descent, with the batch size chosen to fit the entire training data. However, we found that (i)~large batch sizes are detrimental for both the ``normally'' trained models and models trained with reconstructed data, and (ii)~ data reconstruction also works with smaller batch sizes (\emph{e.g.} 64). We thus used mini-batch SGD as the optimizer.

\subsection{Computing the Similarity Between Reconstructed and Reference Samples}\label{sec:similarity_computation}
In this section, we give details on how we compute the similarity between reconstructed and reference samples used in the proposed \emph{supervised} strategy (see \Cref{sec:hyperparameter_tuning_strategies}). We also suggest two possible improvements that could be explored in future works.

For computing image similarity, we closely follow \citep{haim2022reconstructing} in performing the following two steps.

\textbf{Step 1: Distance calculation.} For each reference sample $x_{\text{ref}}$, we compute the following distance to all reconstructed samples $x_{\text{recon}}$:
\begin{equation}
    d(x_{\text{ref}}, x_{\text{recon}}) = \left\| \hat{x}_{\text{ref}} -  \hat{x}_{\text{recon}} \right\|_{2}^{2},
\end{equation}
where $\hat{x}_{\text{ref}}$ is obtained by reducing from $x_{\text{ref}}$ the mean of all reference samples and then dividing it by the standard deviation of all reference samples. The same steps are respectively applied to obtain $\hat{x}_{\text{recon}}$. We select for each training sample its closest reconstructed neighbour.

\textbf{Step 2: Sample creation.}: (i)~ We re-add the training set mean (note that our data is normalized to $[-1, 1]$ by reducing the training mean from all data points) to reconstructed samples $\hat{x}_{\text{recon}}$ and reference samples $x_{\text{ref}}$. (ii)~We then stretch the  $x_{\text{ref}}$ to $[0,1]$ by scaling according to the minimum and maximum value. (iii)~Lastly, we compute the SSIM \citep{wang2004image} between the pair.

\textbf{Possible improvements.} We see the following possible ways to improve the reconstruction process. (i) We remark that assessing image similarity (or aesthetics of generated images) is a subjective and open process in research. However, we think that future work could improve our \frameworkshort by filtering for reconstructed samples that fill certain \emph{real data-like} criteria. Previous works by \citep{haim2022reconstructing} used a SSIM score of above $0.4$ as threshold, but other thresholds and similarity metrics are possible. 

(ii) We currently do not enforce reconstructed samples to be unique. As a result, two reconstructed samples might provide the same (non-)new information to the machine learning model. Here, combining close reconstructed samples by, \emph{e.g.}, na{\"i}ve overlaying or more sophisticated dataset distillation techniques \citep[cf.][]{loo2023understanding,yang2023efficient}. These methods are applicable post-reconstruction. Alternatively one could enforce pairwise dissimilarity during minimization of \Cref{eq:loss_full}.

\clearpage
\section{Result for CIL}\label{sec:appendix_results_cil}
In this section, we provide extended results for the experiments reported in \Cref{sec:main_cil}. \Cref{tab:cil_varying_reconstruction_epochs} shows ACC and BWT for a MLP trained with varying reconstructed samples on SplitMNIST.

\subsection{Sensitivity Analysis for the CIL scenario}\label{sec:appendix_results_cil_sensitivity}
In this section, we give details on the sensitivity analysis of our \frameworkshort to the number of reconstruction samples $m$. Results are in \Cref{tab:class_sensitivity}. \frameworkshort is already competitive when merely reconstructing $m=10$ training samples.
\begin{table}[h!]
\footnotesize
\centering
\caption{\textbf{Scenario CIL:} Classification performance with varying number of reconstruction samples $m$. Shown: average ACC[$\uparrow$] and BWT[$\uparrow$] over 5 random repetitions with varying task order and initialization.}
\label{tab:class_sensitivity}
\begin{tabular}{llS[table-format=2.2(1)]S[table-format=2.2(4)]}
\toprule
{$m$} & \textbf{Method} & {ACC(±std)} & {BWT(±std)} \\
\midrule
\multirow{3}{*}{10}
 & Finetune+\frameworkshort & 49.36(4.13) & -51.02(5.63) \\
 & Finetune+\frameworkshort (unsup.) & 61.92(3.73) & -36.16(8.25) \\
 & Finetune+\frameworkshort (sup.) & 54.34(5.66) & -42.14(12.08) \\
\midrule
\multirow{3}{*}{25}
 & Finetune+\frameworkshort & 58.16(0.34) & -38.58(1.28) \\
 & Finetune+\frameworkshort (unsup.) & 60.54(2.62) & -38.07(5.58) \\
 & Finetune+\frameworkshort (sup.) & 56.32(3.83) & -46.04(4.88) \\
\midrule
\multirow{3}{*}{50}
 & Finetune+\frameworkshort & 63.83(3.69) & -31.91(5.44) \\
 & Finetune+\frameworkshort (unsup.) & 61.39(7.28) & -34.13(9.13) \\
 & Finetune+\frameworkshort (sup.) & 59.37(5.50) & -39.15(7.42) \\
\midrule
\multirow{3}{*}{75}
 & Finetune+\frameworkshort & 73.41(1.38) & -20.67(1.58) \\
 & Finetune+\frameworkshort (unsup.) & 64.24(2.02) & -25.03(10.42) \\
 & Finetune+\frameworkshort (sup.) & 67.07(4.09) & -24.26(10.61) \\
\midrule
\multirow{3}{*}{85}
 & Finetune+\frameworkshort & 67.30(3.07) & -28.25(4.42) \\
 & Finetune+\frameworkshort (unsup.) & 67.13(3.11) & -16.36(7.76) \\
 & Finetune+\frameworkshort (sup.) & 64.84(5.88) & -29.69(9.93) \\
\midrule
\multirow{3}{*}{100}
 & Finetune+\frameworkshort & 71.76(1.88) & -21.82(1.34) \\
 & Finetune+\frameworkshort (unsup.) & 66.99(3.05) & -20.64(9.20) \\
 & Finetune+\frameworkshort (sup.) & 67.45(3.17) & -23.49(7.81) \\
\midrule
\multirow{3}{*}{110}
 & Finetune+\frameworkshort & 70.18(0.84) & -22.92(0.84) \\
 & Finetune+\frameworkshort (unsup.) & 66.66(2.48) & -13.55(2.78) \\
 & Finetune+\frameworkshort (sup.) & 69.99(0.72) & -14.51(5.44) \\
\midrule
\multirow{3}{*}{125}
 & Finetune+\frameworkshort & 65.08(1.18) & -29.24(3.22) \\
 & Finetune+\frameworkshort (unsup.) & 62.11(5.51) & -13.87(6.15) \\
 & Finetune+\frameworkshort (sup.) & 67.84(4.59) & -18.87(8.19) \\
\bottomrule
\end{tabular}
\end{table}
\vfill\eject
\subsection{CIFAR10 visualization}
Below, we provide visualizations for \frameworkshort with state-of-the-art CL methods on the CIFAR10 dataset. Tabular results are in \Cref{tab:CIL_mlp_main} in the main text, \Cref{sec:main_cil}.
\begin{figure}[!hbtp]
\includegraphics[width=\linewidth]{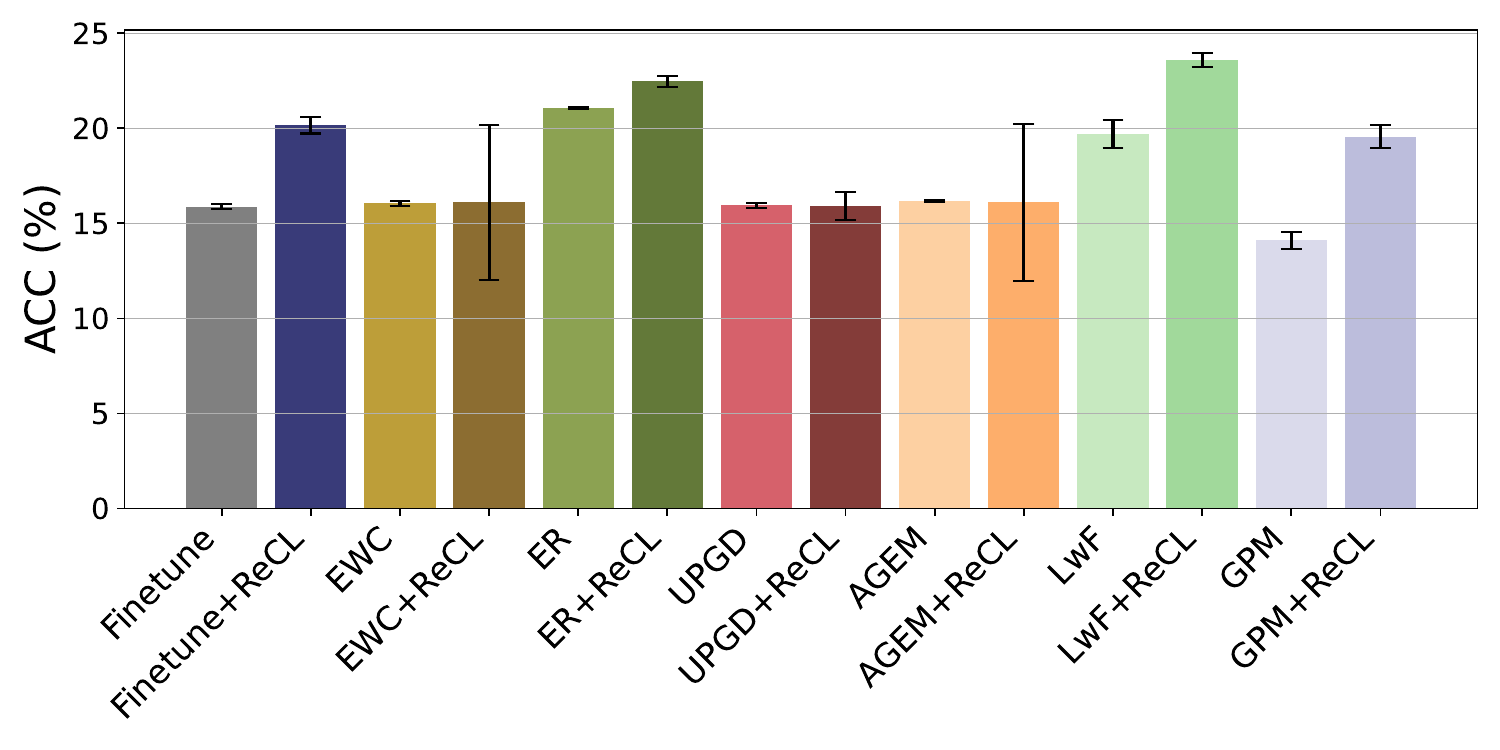}
     \caption{\textbf{ACC [$\uparrow$] for CIL,  SplitCIFAR10.} All methods benefit from our \frameworkshort. Used standalone, \frameworkshort already is competitive to CL methods.}
     \label{fig:cil_acc_cifar}
\end{figure}
\begin{figure}[!hbtp]
\includegraphics[width=\linewidth]{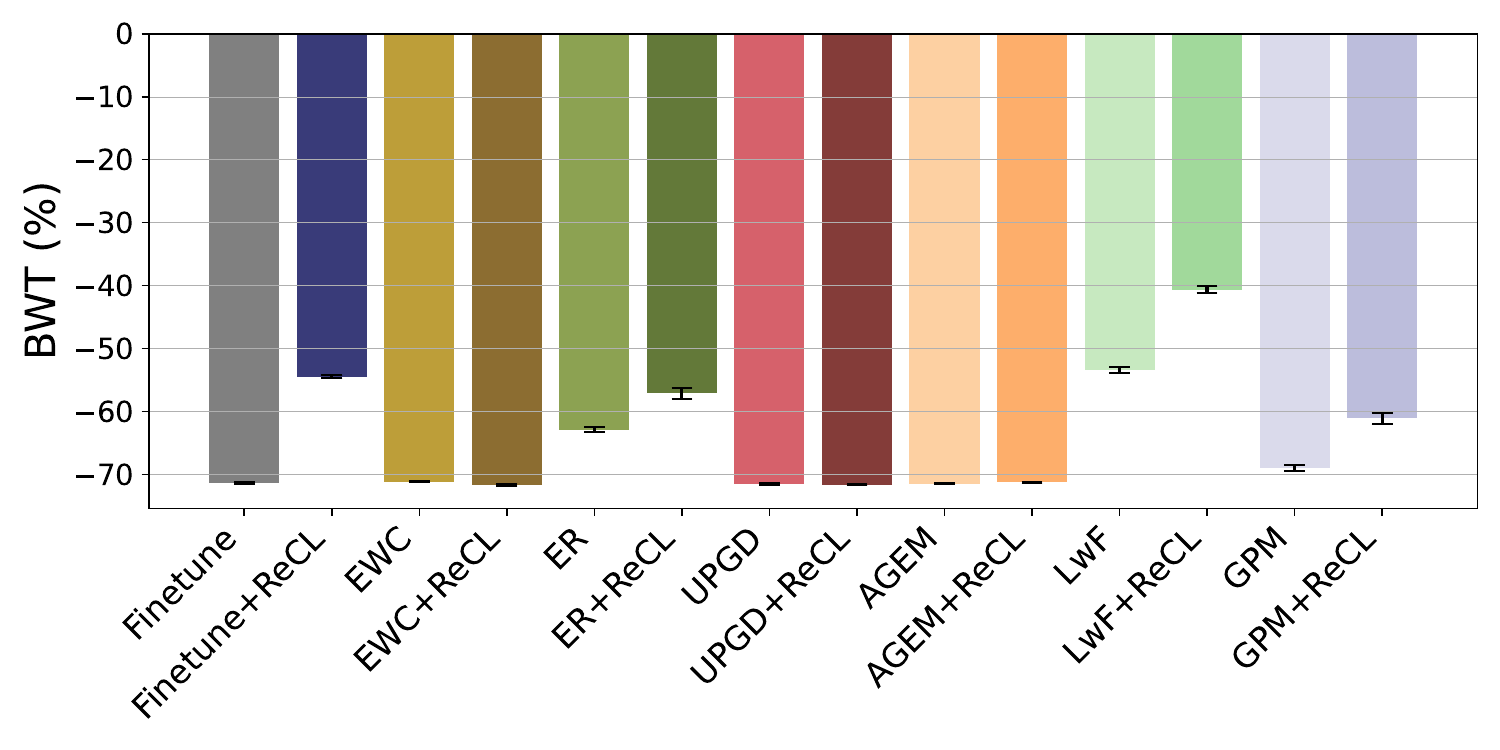}
          \caption{\textbf{BWT [$\uparrow$] for CIL, SplitCIFAR10.} \frameworkshort strongly reduces forgetting for all methods and is competitive when used standalone.}
     \label{fig:cil_bwt_cifar}
\end{figure}

\clearpage
\section{Results for DIL}\label{sec:appendix_results_dil}
In this section, we provide the full results for the experiments reported in \Cref{sec:main_dil}. \Cref{tab:cil_varying_reconstruction_epochs} shows ACC and BWT for a MLP trained with varying reconstructed samples on SplitMNIST.

\subsection{Sensitivity Analysis DIL}\label{sec:appendix_dil_sensitivity}
In this section, we give details on the sensitivity analysis of our \frameworkshort to the number of reconstruction samples $m$ in the DIL scenario. The results are in \Cref{tab:domain_sensitivity}. \frameworkshort is already competitive when merely reconstructing $m=10$ samples while adding minimal computational overhead.
\begin{table}[h!]
\footnotesize
\centering
\caption{\textbf{Scenario DIL:} Classification performance with varying number of reconstruction samples $m$. Shown: average ACC[$\uparrow$] and BWT[$\uparrow$] over 5 random repetitions with varying task order and initialization.}
\label{tab:domain_sensitivity}
\begin{tabular}{llS[table-format=2.2(2)]S[table-format=2.2(2)]}
\toprule
{$m$} & \textbf{Method} & {ACC(±std)} & {BWT(±std)} \\
\midrule
\multirow{3}{*}{10}
 & Finetune+ReCL & 80.17(0.13) & -9.64(2.25) \\
 & Finetune+ReCL (unsup.) & 81.19(0.71) & -8.35(1.01) \\
 & Finetune+ReCL (sup.) & 80.99(0.76) & -10.64(2.37) \\
\midrule
\multirow{3}{*}{25}
 & Finetune+ReCL & 81.58(0.31) & -7.74(2.20) \\
 & Finetune+ReCL (unsup.) & 81.39(0.23) & -7.67(3.79) \\
 & Finetune+ReCL (sup.) & 80.98(0.81) & -9.48(2.95) \\
\midrule
\multirow{3}{*}{50}
 & Finetune+ReCL & 81.75(0.40) & -7.64(1.26) \\
 & Finetune+ReCL (unsup.) & 81.18(0.25) & -7.91(0.55) \\
 & Finetune+ReCL (sup.) & 80.91(0.85) & -10.23(2.57) \\
\midrule
\multirow{3}{*}{75}
 & Finetune+ReCL & 79.74(1.46) & -17.87(1.81) \\
 & Finetune+ReCL (unsup.) & 81.72(0.55) & -4.55(0.93) \\
 & Finetune+ReCL (sup.) & 81.17(0.71) & -9.42(1.48) \\
\midrule
\multirow{3}{*}{85}
 & Finetune+ReCL & 80.83(0.08) & -10.33(2.69) \\
 & Finetune+ReCL (unsup.) & 81.51(0.39) & -5.29(1.34) \\
 & Finetune+ReCL (sup.) & 81.19(0.77) & -9.56(2.61) \\
\midrule
\multirow{3}{*}{100}
& Finetune+\frameworkshort & 83.80(0.06) & -4.91(0.13) \\
 & Finetune+ReCL (unsup.) & 81.36(0.64) & -5.69(2.39) \\
 & Finetune+ReCL (sup.) & 80.84(0.63) & -10.33(2.66) \\
\midrule
\multirow{3}{*}{110}
 & Finetune+ReCL & 79.22(2.24) & -20.02(2.62) \\
 & Finetune+ReCL (unsup.) & 81.82(0.17) & -8.10(2.76) \\
 & Finetune+ReCL (sup.) & 81.07(0.50) & -8.79(0.42) \\
\midrule
\multirow{3}{*}{125}
 & Finetune+ReCL & 80.39(2.44) & -17.97(3.20) \\
 & Finetune+ReCL (unsup.) & 81.35(0.32) & -6.18(1.47) \\
 & Finetune+ReCL (sup.) & 80.82(0.62) & -10.79(2.01) \\
\bottomrule
\end{tabular}
\end{table}

\subsection{CIFAR10 visualization}
We here provide visualizations for \frameworkshort with state-of-the-art CL methods on the CIFAR10 dataset. Tabular results are in \Cref{tab:DIL_mlp_main} in the main text, \Cref{sec:main_dil}.
\begin{figure}[!hbtp]
\includegraphics[width=\linewidth]{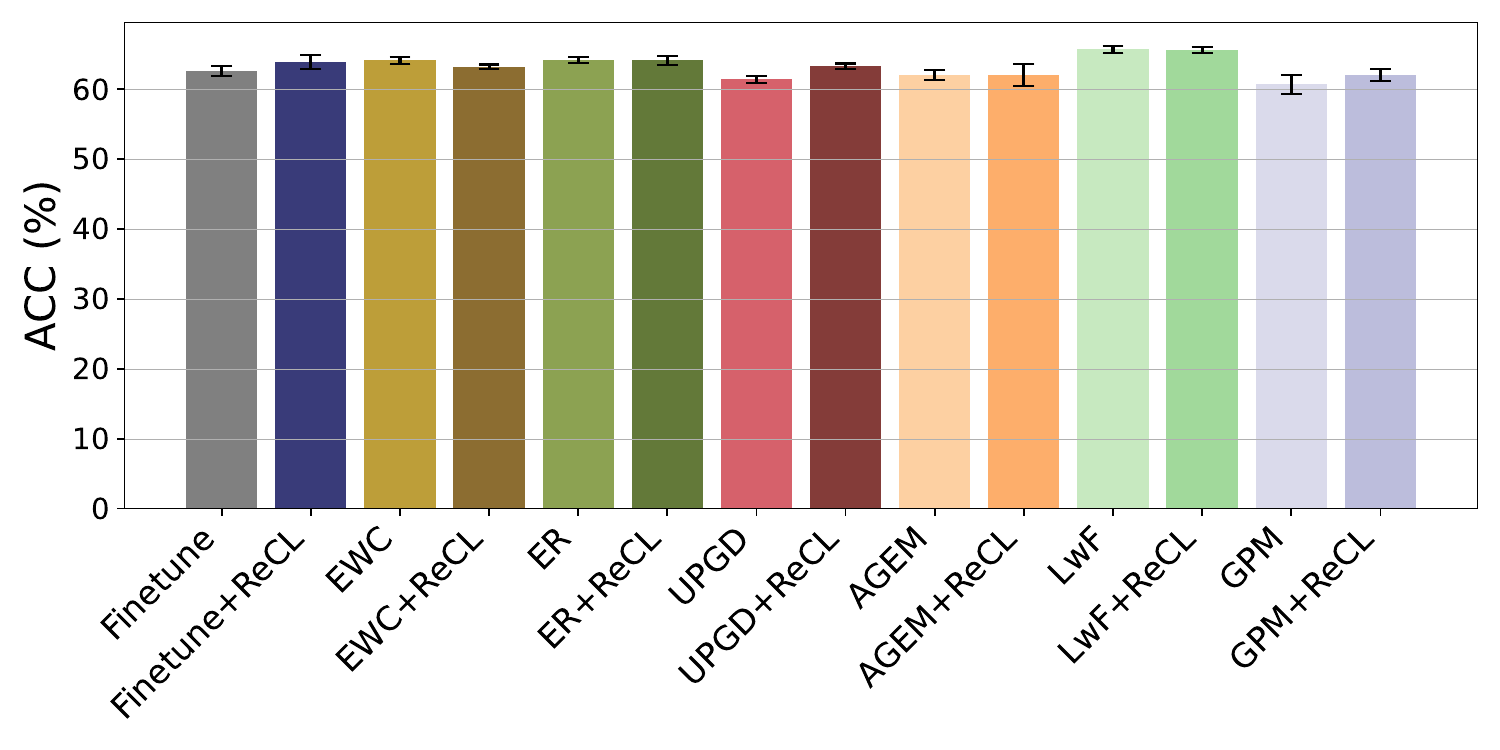}
     \caption{\textbf{ACC [$\uparrow$] for DIL, SplitCIFAR10.} All methods benefit from our \frameworkshort. Used standalone, \frameworkshort already is competitive to CL methods.}
     \label{fig:dil_acc_cifar}
\end{figure}
\begin{figure}[!hbtp]
\includegraphics[width=\linewidth]{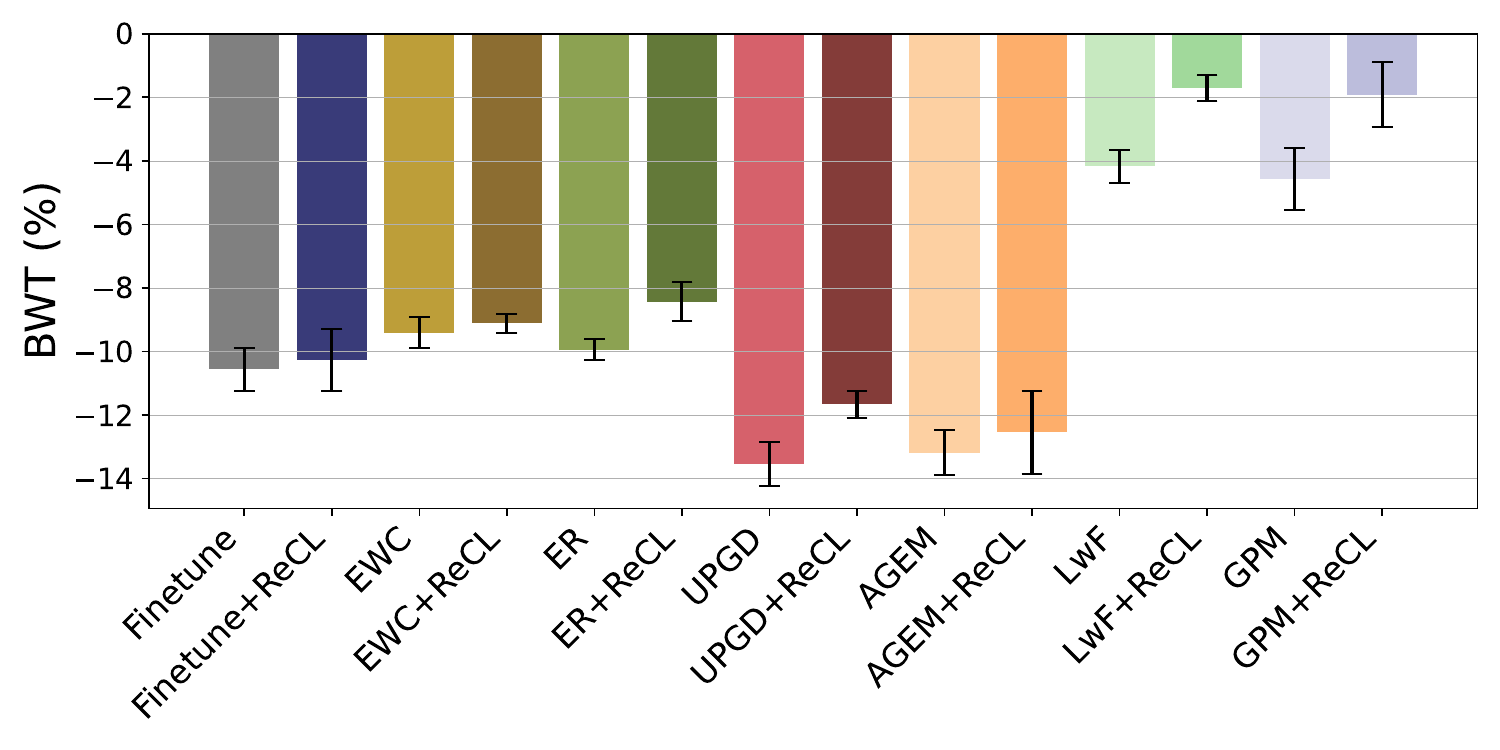}
          \caption{\textbf{BWT [$\uparrow$] for DIL, SplitCIFAR10.} \frameworkshort strongly reduces forgetting for all methods and is competitive when used standalone.}
     \label{fig:dil_bwt_cifar}
\end{figure}

\clearpage
\section{Results on TinyImageNet}
\label{app:tinyimagenet}
\begin{figure}[!hbtp]
\includegraphics[width=\linewidth]{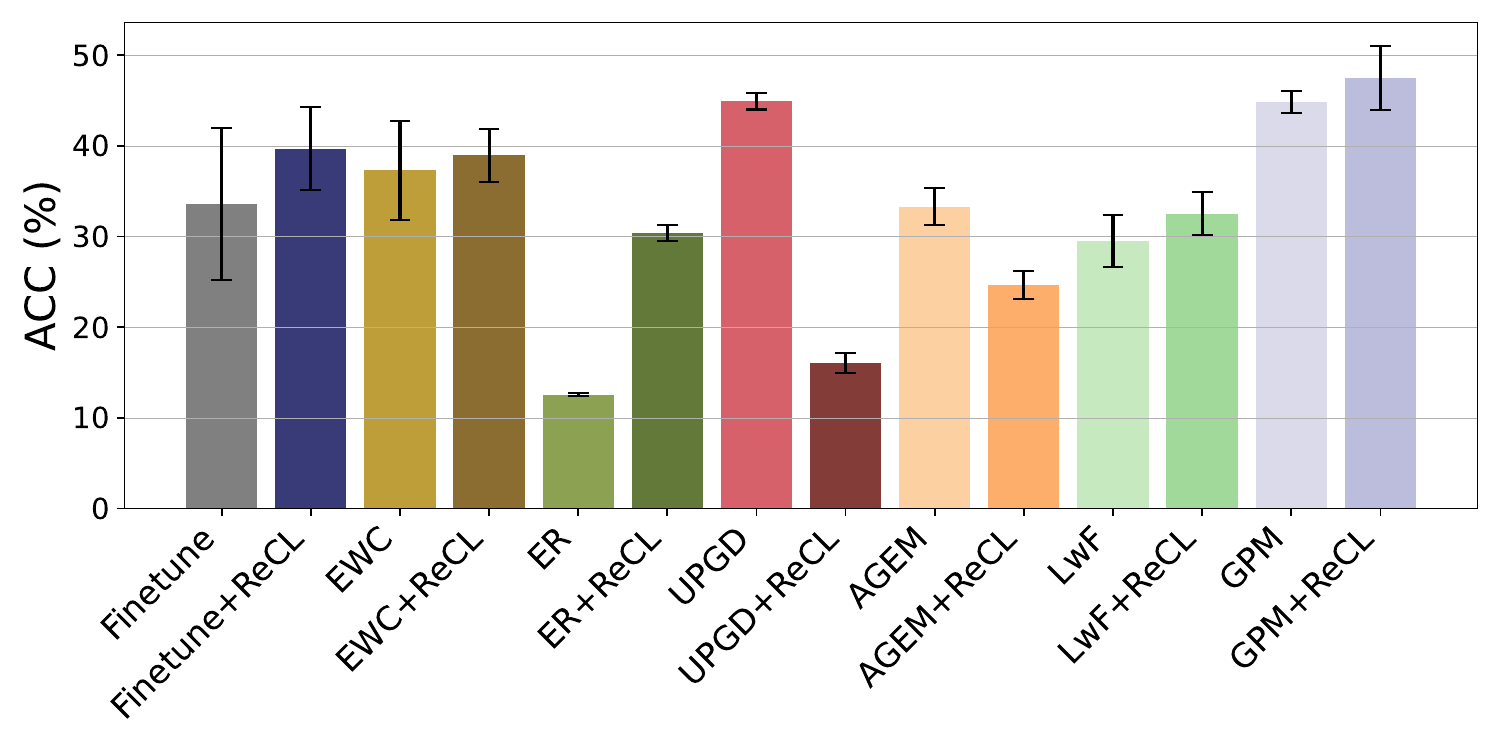}
     \caption{\textbf{ACC [$\uparrow$] for CIL, SplitTinyImageNet.} All methods benefit from our \frameworkshort. Used standalone, \frameworkshort already is competitive to CL methods.}
     \label{fig:tinyimagenet_class_appendix}
\end{figure}

\begin{figure}[!hbtp]
\includegraphics[width=\linewidth]{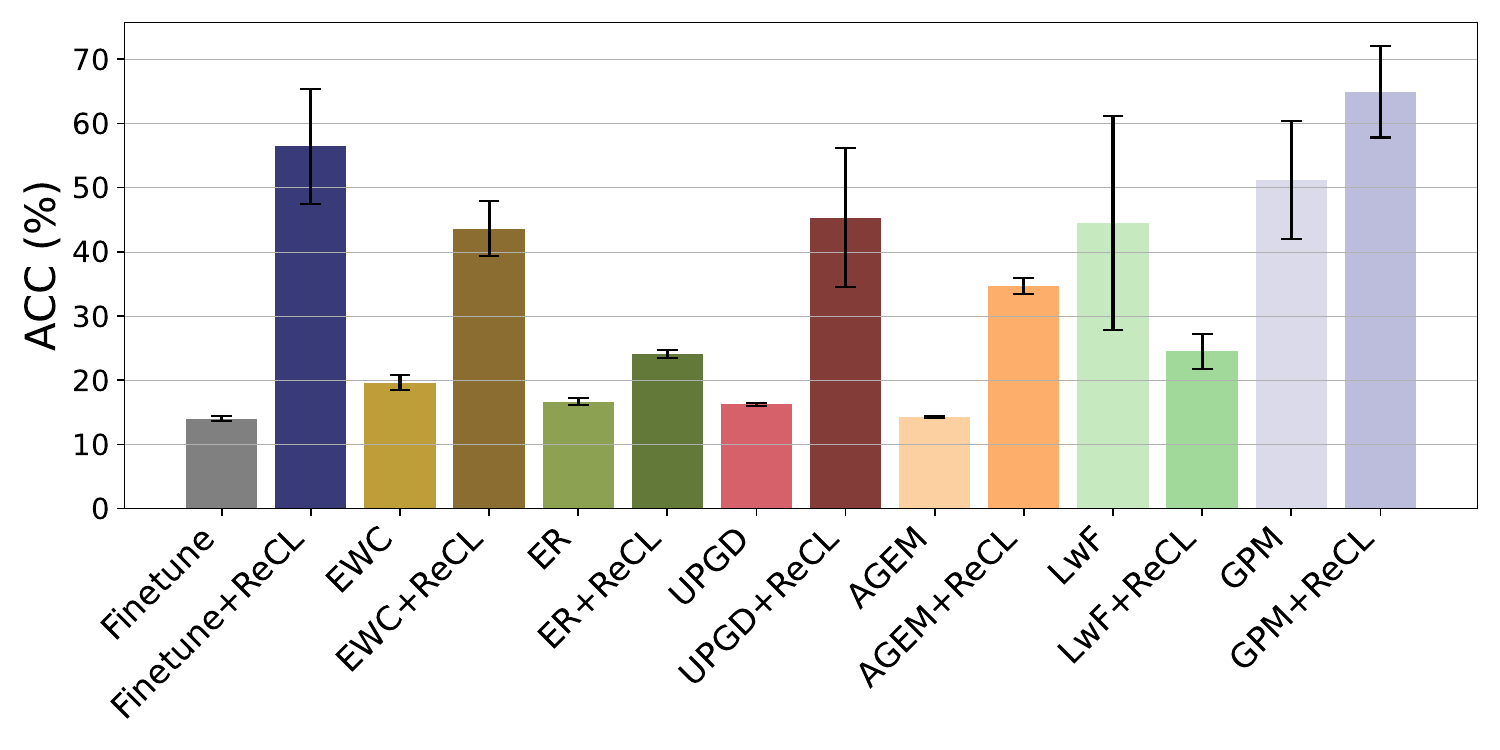}
     \caption{\textbf{ACC [$\uparrow$] for DIL, SplitTinyImageNet.} All methods benefit from our \frameworkshort. Used standalone, \frameworkshort already is competitive to CL methods.}
     \label{fig:tinyimagenet_domain_appendix}
\end{figure}

In this section, we provide the full results for the experiments reported in \Cref{sec:tinyimagenet}.  \Cref{tab:tinyimagenet_cil,tab:tinyimagenet_dil} show the tabular results for \Cref{fig:tinyimagenet_class_appendix,fig:tinyimagenet_domain_appendix}.
\begin{table}[ht]
\centering
\caption{\textbf{Scenario CIL:} Results on SplitTinyImageNet. Shown: average ACC[$\uparrow$] and BWT[$\uparrow$] over 5 random repetitions with varying order and init.}
\label{tab:tinyimagenet_cil}
\begin{tabular}{lS[table-format=2.2(2)]S[table-format=3.2(4)]}
\toprule
Approach & {ACC(±std)} & {BWT(±std)} \\
\midrule
Finetune & 33.58(8.40) & -47.50(23.52)\\
Finetune+\frameworkshort & 39.68(4.56) & -35.00(0.59) \\
\midrule
ewc & 37.28(5.49) & -79.20(7.07) \\
ewc+\frameworkshort & 38.93(2.89) & -33.77(2.68) \\
\midrule
er & 12.57(0.15) & -24.20(0.71) \\
er+\frameworkshort & 30.38(0.84) & -50.57(1.25) \\
\midrule
upgd & 44.93(0.94) & -88.73(3.18) \\
upgd+\frameworkshort & 16.05(1.13) & -36.73(4.53) \\
\midrule
agem & 33.28(2.03) & -66.93(7.33) \\
agem+\frameworkshort & 24.62(1.55) & -48.47(0.88) \\
\midrule
lwf & 29.53(2.86) & -17.80(6.76) \\
lwf+\frameworkshort & 32.53(2.40) & -17.80(4.85) \\
\midrule
gpm & 44.82(1.20) & -87.17(2.44) \\
gpm+\frameworkshort & 47.48(3.56) & -21.03(10.72) \\
\bottomrule
\end{tabular}
\end{table}
\begin{table}[ht]
\centering
\caption{\textbf{Scenario DIL:} Results on SplitTinyImageNet. Shown: average ACC[$\uparrow$] and BWT[$\uparrow$] over 5 random repetitions with varying order and init.}
\label{tab:tinyimagenet_dil}
\begin{tabular}{lS[table-format=2.2(4)]S[table-format=3.2(2)]}
\toprule
Approach & {ACC(±std)} & {BWT(±std)} \\
\midrule
Finetune & 14.03(0.41) & -19.90(3.05) \\
Finetune+\frameworkshort & 56.42(8.96) & -33.13(1.92) \\
\midrule
EWC & 19.63(1.23) & -17.07(0.66) \\
EWC+\frameworkshort & 43.62(4.27) & -31.27(2.49) \\
\midrule
ER & 16.67(0.50) & -11.73(0.90) \\
ER+\frameworkshort & 24.07(0.67) & -14.67(0.21) \\
\midrule
UPGD & 16.22(0.19) & -17.13(0.21) \\
UPGD+\frameworkshort & 45.27(10.82) & -49.93(6.08) \\
\midrule
AGEM & 14.28(0.18) & -17.60(0.14) \\
AGEM+\frameworkshort & 34.65(1.21) & -23.87(0.68) \\
\midrule
LwF & 44.43(16.67) & -7.07(1.93) \\
LwF+\frameworkshort & 24.50(2.72) & -16.70(0.24) \\
\midrule
GPM & 51.20(9.22) & -27.27(2.47) \\
GPM+\frameworkshort & 64.95(7.15) & -15.90(1.79) \\
\bottomrule
\end{tabular}
\end{table}

\clearpage
\section{Varying the number of reconstruction epochs}\label{sec:varying_reconstruction_epochs}
Here, we present results when varying the number of reconstruction epochs in the CIL and DIL scenario. We train a MLP on the SplitMNIST dataset and vary {$n_{\text{rec}}$} as $[500, 1000, 2500, 5000, 10000]$.

The results are in \Cref{tab:cil_varying_reconstruction_epochs,tab:dil_varying_reconstruction_epochs}. We observe that more reconstruction epochs are mostly beneficial in the CIL scenario, whereas they have little effect on the DIL scenario.

\begin{table}[!htbp]
\footnotesize
\centering
\caption{\textbf{Scenario CIL:} Results for varying reconstruction epochs $n_{\text{rec}}$ (default: $1000$). Shown: average ACC[$\uparrow$], BWT[$\uparrow$], and runtime over 5 random repetitions with varying task order and initialization.}
\label{tab:cil_varying_reconstruction_epochs}
\begin{tabular}{llS[table-format=2.2(2)]S[table-format=2.2(4)]}
\toprule
{$m$} & \textbf{Method} & {ACC(±std)} & {BWT(±std)} \\
\midrule
\multirow{1}{*}{500}
 & Finetune+\frameworkshort & 70.04(3.06) & -23.38(7.29) \\
\midrule
\multirow{1}{*}{1000}
& Finetune+\frameworkshort & 73.18(0.09) & -14.51(1.64) \\
\midrule
\multirow{1}{*}{2500}
 & Finetune+\frameworkshort & 68.80(1.25) & -22.36(4.28) \\
\midrule
\multirow{1}{*}{5000}
 & Finetune+\frameworkshort & 66.94(1.19) & -22.69(10.05) \\
\midrule
\multirow{1}{*}{10000}
 & Finetune+\frameworkshort & 67.87(3.65) & -23.93(8.60) \\
\bottomrule
\end{tabular}
\end{table}

\begin{table}[!htbp]
\footnotesize
\centering
\caption{\textbf{Scenario DIL:} Results for varying reconstruction epochs $n_{\text{rec}}$ (default: $1000$). Shown: average ACC[$\uparrow$], BWT[$\uparrow$], and runtime over 5 random repetitions with varying task order and initialization.}
\label{tab:dil_varying_reconstruction_epochs}
\begin{tabular}{llS[table-format=2.2(2)]S[table-format=2.2(2)]}
\toprule
{$m$} & \textbf{Method} & {ACC(±std)} & {BWT(±std)} \\
\midrule
\multirow{1}{*}{500}
 & Finetune+ReCL & 73.84(3.96) & 4.20(3.38) \\
\midrule
\multirow{1}{*}{1000}
 & Finetune+\frameworkshort & 83.80(0.06) & -4.91(0.13) \\
\midrule
\multirow{1}{*}{2500}
 & Finetune+ReCL & 78.05(0.46) & -1.43(3.39) \\
\midrule
\multirow{1}{*}{5000}
 & Finetune+ReCL & 73.42(6.20) & 0.82(4.85) \\
\midrule
\multirow{1}{*}{10000}
 & Finetune+ReCL & 74.47(5.95) & 0.08(4.16) \\
\bottomrule
\end{tabular}
\end{table}

\clearpage
\section{Extended results for large-scale datasets}\label{sec:scaling_recl}
In this section, we present extended results for applying \frameworkshort to large-scale (i.e., full) datasets. For this, we select SplitMNIST and SplitTinyImageNet and use the entire training data. We use the same architectures (\Cref{app:imp_details}) as before, 
but set the number of reconstruction samples to \num{1000} and \num{600} for SplitMNIST and SplitTinyImageNet, respectively. Again, we optimize the hyperparameters for each experiment separately over the grid given in \Cref{app:imp_details,tab:table_search_space}. The results are in \Cref{tab:cil_mnist_full} to \Cref{tab:dil_imgnt_full}.

Our findings align with our previous observations, and \frameworkshort also slows down forgetting for larger datasets. Note how \frameworkshort can even lead to retrospective performance improvements on past tasks, as the \emph{positive} BWT for method LwF indicates.

\begin{table}[ht]
\centering
\caption{\textbf{Scenario CIL: Results on \emph{full} SplitMNIST.} Shown: average ACC[$\uparrow$] and BWT[$\uparrow$] over 5 random repetitions with varying task order and init.}
\label{tab:cil_mnist_full}
\begin{tabular}{lS[table-format=2.2(2)]S[table-format=3.2(2)]}
\toprule
Approach & {ACC(±std)} & {BWT(±std)} \\
\midrule
Finetune & 20.08(0.21) & -99.45(0.40) \\
Finetune+\frameworkshort & 57.74(0.36) & -50.49(3.62) \\
\midrule
EWC & 20.35(0.53) & -99.05(0.81) \\
EWC+\frameworkshort & 61.50(1.16) & -44.54(2.64) \\
\midrule
ER & 93.73(1.24) & -6.98(0.38) \\
ER+\frameworkshort & 93.83(1.14) & -6.55(0.54) \\
\midrule
UPGD & 56.65(0.40) & -45.52(11.52) \\
UPGD+\frameworkshort & 59.20(1.65) & -47.61(7.49) \\
\midrule
AGEM & 20.01(0.11) & -99.64(0.14) \\
AGEM+\frameworkshort & 66.68(0.68) & -38.60(4.39) \\
\midrule
LwF & 45.35(1.90) & -39.55(6.29) \\
LwF+\frameworkshort & 59.08(0.27) & 4.58(2.33) \\
\midrule
GPM & 19.87(0.23) & -98.47(0.78) \\
GPM+\frameworkshort & 55.97(1.03) & -51.77(7.20) \\
\bottomrule
\end{tabular}
\end{table}
\begin{table}[ht]
\centering
\caption{\textbf{Scenario CIL: Results on \emph{full} SplitTinyImageNet.} Shown: average ACC[$\uparrow$] and BWT[$\uparrow$] over 5 random repetitions with varying task order and init.}
\label{tab:cil_imgnt_full}
\begin{tabular}{lS[table-format=2.2(2)]S[table-format=3.2(2)]}
\toprule
Approach & {ACC(±std)} & {BWT(±std)} \\
\midrule
Finetune & 29.85(3.44) & -51.23(4.78) \\
Finetune+\frameworkshort & 28.85(4.19) & -46.70(3.41) \\
\midrule
EWC & 33.10(4.65) & -60.70(11.89) \\
EWC+\frameworkshort & 37.55(3.72) & -50.60(2.87) \\
\midrule
ER & 33.33(2.01) & -58.23(4.45) \\
ER+\frameworkshort & 32.17(2.71) & -45.47(3.76) \\
\midrule
UPGD & 35.42(4.36) & -69.10(9.41) \\
UPGD+\frameworkshort & 26.45(4.51) & -47.80(16.23) \\
\midrule
AGEM & 20.10(0.97) & -48.07(1.59) \\
AGEM+\frameworkshort & 27.38(1.16) & -49.43(0.75) \\
\midrule
LwF & 44.68(1.64) & -26.43(4.01) \\
LwF+\frameworkshort & 65.63(4.43) & -19.10(4.09) \\
\midrule
GPM & 28.08(2.34) & -67.17(6.18) \\
GPM+\frameworkshort & 40.55(2.87) & -49.83(6.96) \\
\bottomrule
\end{tabular}
\end{table}
\begin{table}[ht]
\centering
\caption{\textbf{Scenario DIL: Results on \emph{full} SplitMNIST.} Shown: average ACC[$\uparrow$] and BWT[$\uparrow$] over 5 random repetitions with varying task order and init.}
\label{tab:dil_mnist_full}
\begin{tabular}{lS[table-format=2.2(2)]S[table-format=3.2(2)]}
\toprule
Approach & {ACC(±std)} & {BWT(±std)} \\
\midrule
Finetune & 79.26(1.01) & -16.78(3.03) \\
Finetune+\frameworkshort & 73.59(6.88) & -32.44(8.03) \\
\midrule
EWC & 74.79(7.76) & -30.90(9.23) \\
EWC+\frameworkshort & 76.49(0.96) & -22.73(1.67) \\
\midrule
ER & 96.85(0.15) & -3.19(0.18) \\
ER+\frameworkshort & 96.44(0.33) & -3.62(0.32) \\
\midrule
UPGD & 76.94(0.73) & -21.62(1.96) \\
UPGD+\frameworkshort & 78.99(0.89) & -17.74(2.80) \\
\midrule
AGEM & 73.80(6.45) & -32.30(8.25) \\
AGEM+\frameworkshort & 77.18(6.23) & -27.85(7.26) \\
\midrule
LwF & 83.27(1.73) & -1.17(1.64) \\
LwF+\frameworkshort & 82.58(2.36) & 8.58(0.95) \\
\midrule
GPM & 66.56(3.08) & -0.73(0.90) \\
GPM+\frameworkshort & 76.28(4.25) & -28.22(4.54) \\
\bottomrule
\end{tabular}
\end{table}

\begin{table}[ht]
\centering
\caption{\textbf{Scenario DIL: Results on \emph{full} SplitTinyImageNet.} Shown: average ACC[$\uparrow$] and BWT[$\uparrow$] over 5 random repetitions with varying task order and init.}
\label{tab:dil_imgnt_full}
\begin{tabular}{lS[table-format=2.2(4)]S[table-format=3.2(2)]}
\toprule
Approach & {ACC(±std)} & {BWT(±std)} \\
\midrule
Finetune & 54.80(8.80) & -40.17(2.23) \\
Finetune+\frameworkshort & 23.77(4.78) & -12.20(4.74) \\
\midrule
EWC & 45.68(9.66) & -40.30(3.77) \\
EWC+\frameworkshort & 25.78(2.10) & -15.30(1.06) \\
\midrule
ER & 44.45(12.41) & -32.53(6.89) \\
ER+\frameworkshort & 44.72(1.90) & -18.40(1.00) \\
\midrule
UPGD & 47.37(4.52) & -52.67(2.41) \\
UPGD+\frameworkshort & 30.38(1.53) & -29.77(2.53) \\
\midrule
AGEM & 41.62(7.71) & -48.03(2.73) \\
AGEM+\frameworkshort & 44.08(2.36) & -35.37(0.97) \\
\midrule
LwF & 54.08(4.61) & -18.50(0.41) \\
LwF+\frameworkshort & 63.22(20.13) & -12.77(3.76) \\
\midrule
GPM & 57.05(14.09) & -36.53(1.52) \\
GPM+\frameworkshort & 54.32(2.54) & -12.67(5.00) \\
\bottomrule
\end{tabular}
\end{table}
\clearpage
\section{\frameworkshort slows down forgetting beyond theoretical limitations}\label{sec:alexnet_results}
In the main text, we focus on MLPs and CNNS, which generally conform to the definition of homogenous neural networks (cf. \Cref{app:homogenous_definition}). These type of neural networks converge to margin maximization points \citet{lyu2019gradient,ji2020directional}. To slow down forgetting in CL, our \frameworkshort builds on these convergence points, as they allow the (unsupervised) reconstruction of old training data. Interestingly, \frameworkshort can also slow down forgetting for networks that are explicitly \emph{not} homogenous, such as AlexNet~\cite{krizhevsky2012imagenet} and ResNet~\citep{he2016deep}.

To demonstrate this, we perform the following additional experiment. We optimize the training hyperparameters of AlexNet and Resnet in the DIL scenario, on the SplitCIFAR10 dataset, following the same search procedure as for the other networks (\Cref{sec:experimental_setup} and \Cref{app:imp_details}). That is, over \num{100} hyperparameter trials, we optimize the learning rate and training epochs over the grid details in \Cref{tab:table_search_space}. We then train AlexNet and ResNet with (\textbf{+\frameworkshort}) and without our framework. 

Our results are in \Cref{tab:non_homogenous_app}. We observe: \textbf{\frameworkshort can slow down forgetting for non-homogenous neural networks}. For AlexNet, \frameworkshort improves up to \SI{4.44}{\percent} in terms of ACC, and up to \SI{3}{\percent} for ResNet. These observations are interesting, as the theory underlying our framework is restricted to homogenous neural networks. It is therefore possible that parts of the theory might also extend to non-homogenous architectures. A detailed study is beyond the scope of this paper, and we thus leave it as an interesting direction for future research.
\begin{table}[!h]
\footnotesize
\tabcolsep=0.11cm
\centering
\caption{\textbf{Scenario DIL:} Results for using our framework with AlexNet \citep{krizhevsky2012imagenet}, a \emph{none}-homogenous neural network violating the theoretical background of our framework. We nonetheless see improvements when adding \frameworkshort. Shown: average ACC[$\uparrow$], BWT[$\uparrow$], and runtime over 5 random repetitions with varying task order and initializations. Complements \Cref{tab:alexnet_results} with runtimes.}
\label{tab:non_homogenous_app}
\begin{tabular}{lSS}
\toprule
\textbf{Method} & \multicolumn{2}{c}{\textbf{SplitCIFAR10}}\\
\cmidrule(lr){2-3}
{} & {ACC} & {BWT}\\
\midrule
AlexNet & 56.47 & -31.63  \\
\midrule
AlexNet+\frameworkshort & 59.23 & -28.25 \\
AlexNet+\frameworkshort (unsup.) & 59.04 & -28.24 \\
AlexNet+\frameworkshort (sup.) & 57.03 & -29.26 \\
\midrule
ResNet & 53.53 & -28.17 \\
\midrule
ResNet+\frameworkshort & 55.16 & -29.41\\
ResNet+\frameworkshort (unsup.) & 56.41 & -28.23 \\
ResNet+\frameworkshort (sup.) & 53.51 & -32.69 \\
\bottomrule
\end{tabular}
\end{table}

\section{Scalability Analysis}\label{sec:scalability_analysis}
\begin{figure}
\centering
   \includegraphics[width=0.8\linewidth]{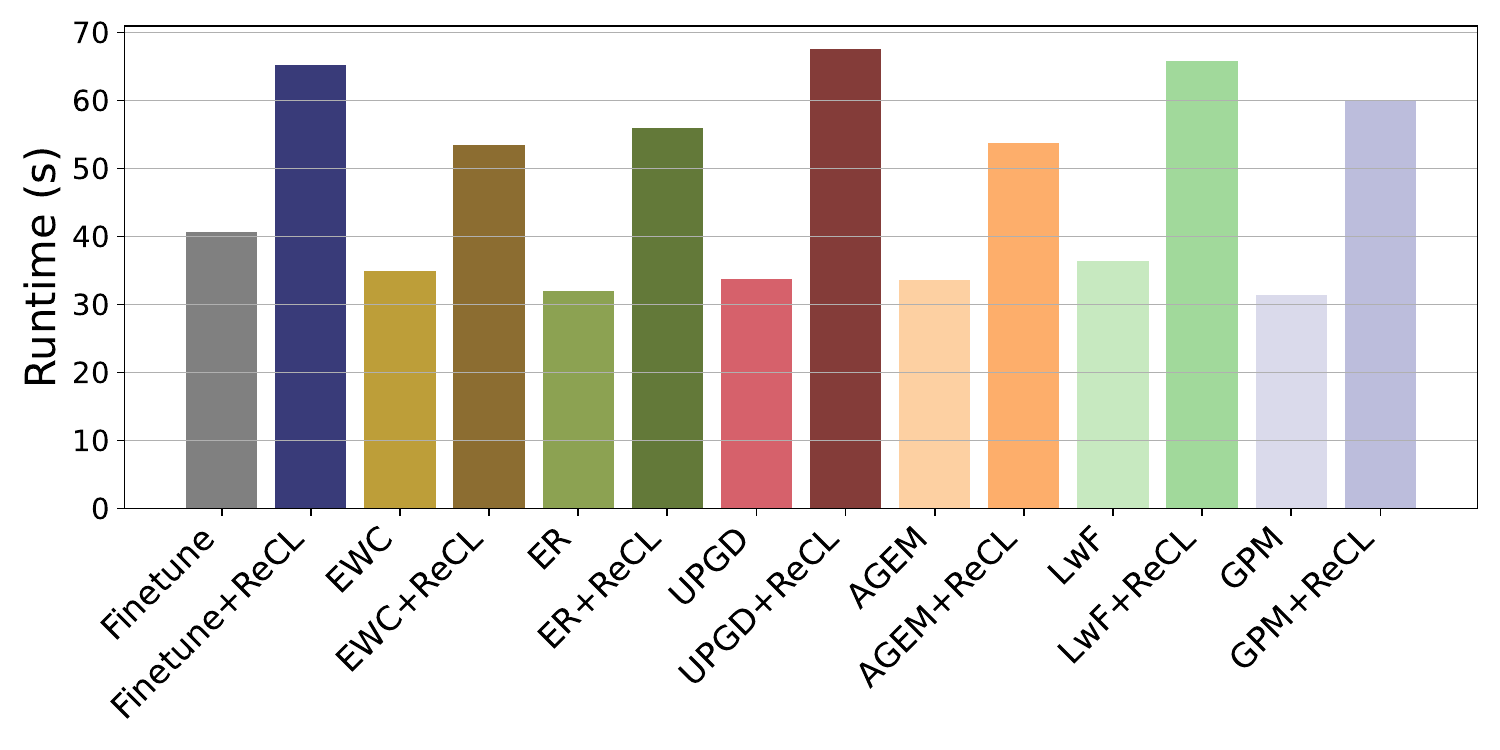}
   \caption{Runtimes for SplitMNIST.}
   \label{fig:runtime_mnist} 
\end{figure}

\begin{figure}
    \centering
    \includegraphics[width=0.8\linewidth]{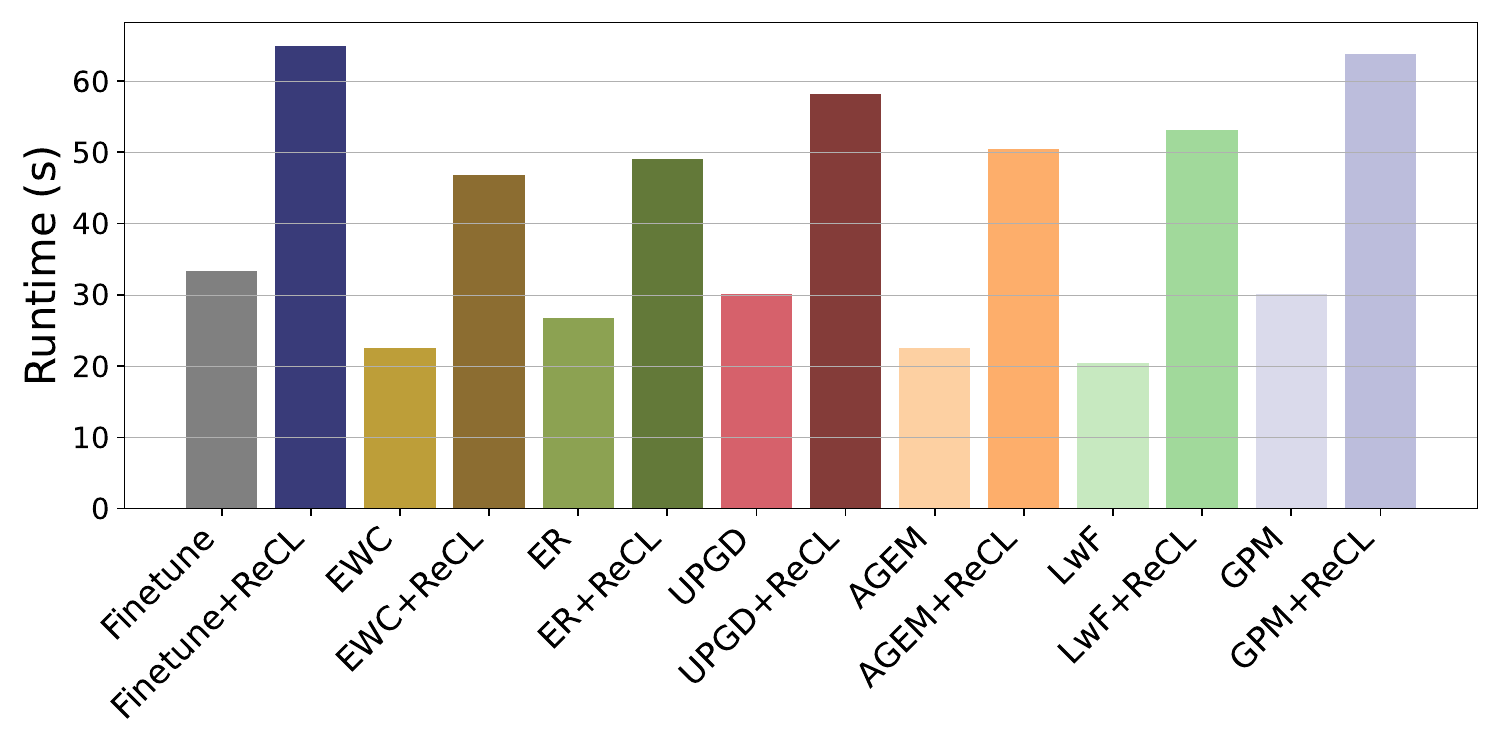}
    \caption{Runtimes for SplitTinyImageNet}
    \label{fig:runtime_tinyimagenet}
\end{figure}

We visualize the runtimes for \frameworkshort in \Cref{fig:runtime_mnist,fig:runtime_tinyimagenet}. We have the following observations: \textbf{(1)}~The overhead for \frameworkshort is relatively small. For example, the runtime of our Finetune+\frameworkshort is roughly twice as long as the Finetune baseline. \textbf{(2)}~~The computational cost is largely insensitive to the dataset type. On SplitMNIST and SplitTinyImageNet, runtime increases are roughly similar. \textbf{(3)}~The runtime of \frameworkshort is insensitive to the underlying CL algorithm. For example, in \Cref{fig:runtime_mnist}, all evaluated CL algorithms show similar runtime increases from our framework. \textbf{(4)} The \emph{in-training} tuning strategies incur larger overheads (cf. \Cref{tab:runtime_analysis}. We thus recommend the na{\"i}ve strategy for use in practice.

\begin{table}[h!]
\tabcolsep=0.11cm
\footnotesize
\centering
\caption{\textbf{Runtime Analysis:} Runtimes on the SplitMNIST and SplitTinyImageNet datasets for the \emph{unsupervised optimization strategy}.}
\label{tab:runtime_analysis}
\begin{tabular}{lSS}
\toprule
\textbf{Approach} & \textbf{MNIST} & \textbf{TinyImageNet} \\
\midrule
Finetune & 40.71 & 33.32 \\
Finetune+ReCL (unsup) & 477.62 & 414.72 \\
\midrule
EWC & 34.93 & 22.60 \\
EWC+ReCL (unsup) & 465.32 & 395.97 \\
\midrule
ER & 31.96 & 26.81 \\
ER+ReCL (unsup) &  459.0 & 397.23 \\
\midrule
UPGD & 33.82 & 30.05 \\
UPGD+ReCL (unsup) & 464.77 & 407.57 \\
\midrule
AGEM & 33.54 & 22.50 \\
AGEM+ReCL (unsup) & 458.40 & 400.91 \\
\midrule
LwF & 36.46 & 20.38 \\
LwF+ReCL (unsup) & 484.21 & 401.90 \\
\bottomrule
\end{tabular}
\end{table}

\end{document}